\documentclass[10pt,twocolumn,letterpaper]{article}
\usepackage{cvpr}              
\usepackage{makecell}
\usepackage{multirow}
\usepackage{xcolor}                 
\definecolor{cvprblue}{rgb}{0.21,0.49,0.74}
\usepackage[pagebackref,breaklinks,colorlinks,allcolors=cvprblue]{hyperref}

\def\paperID{*****} 
\def\confName{CVPR}
\def\confYear{2026}

\title{InEdit-Bench: Benchmarking Intermediate Logical Pathways for\\Intelligent Image Editing Models}

\author{
  Zhiqiang Sheng$^{1,2,3}$, Xumeng Han$^{3,}$\thanks{Project lead.} $^{ , }$\thanks{Corresponding author.} , Zhiwei Zhang$^{1,2,3}$, Zenghui Xiong$^{1,2,3}$, 
   Yifan Ding$^{1,2,3}$,\\ Aoxiang Ping$^{1,2,3}$, Xiang Li$^{1,2,3}$, Tong Guo$^{1,2,3}$,  Yao Mao$^{1,2,3,}$\footnotemark[2] \\
$^{1}$ State Key Laboratory of Optical Field Manipulation Science and Technology,\\ Institute of Optics and Electronics, Chinese Academy of Sciences, Chengdu, 610209, China \\
$^{2}$ National Laboratory on Adaptive Optics, Chengdu, 610209, China\\
$^{3}$ University of Chinese Academy of Sciences, Beijing, 101408, China\\
  \small\texttt{\{shengzhiqiang24, hanxumeng19\}@mails.ucas.ac.cn, maoyao@ioe.ac.cn}\\
  \small \url{https://github.com/SZStrong1/InEdit-Bench}
}

\begin{document}
\maketitle

\begin{figure*}[!ht]
\centering
\includegraphics[width=1\textwidth]{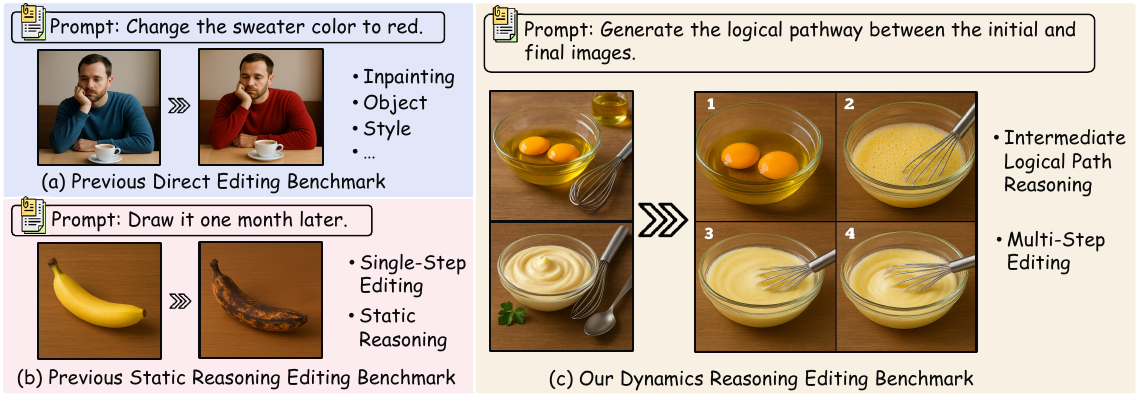} %
\caption{Comparison of previous image editing benchmarks and our proposed InEdit-Bench. (a) Previous Direct Editing Benchmark: Focuses on evaluating the model's ability to execute explicit instructions. (b) Previous Static Reasoning Editing Benchmark: Evaluates the model's ability to perform static reasoning in editing tasks with external knowledge. (c) Our Dynamics Reasoning Editing Benchmark (\textbf{InEdit-Bench}): A benchmark that requires dynamic knowledge reasoning and multi-step planning in editing tasks, aiming to assess the model's comprehensive ability to perform complex, non-direct image editing based on deep semantic understanding.}
\vspace{-12pt}
\label{Instruction_motivation}
\end{figure*}

\vspace{-12pt}
\begin{abstract}
Multimodal generative models have made significant strides in image editing, demonstrating impressive performance on a variety of static tasks. However, their proficiency typically does not extend to complex scenarios requiring dynamic reasoning, leaving them ill-equipped to model the coherent, intermediate logical pathways that constitute a multi-step evolution from an initial state to a final one. This capacity is crucial for unlocking a deeper level of procedural and causal understanding in visual manipulation. To systematically measure this critical limitation, we introduce \textbf{InEdit-Bench}, the first evaluation benchmark dedicated to reasoning over intermediate pathways in image editing. InEdit-Bench comprises meticulously annotated test cases covering four fundamental task categories: state transition, dynamic process, temporal sequence, and scientific simulation. Additionally, to enable fine-grained evaluation, we propose a set of assessment criteria to evaluate the logical coherence and visual naturalness of the generated pathways, as well as the model's fidelity to specified path constraints. Our comprehensive evaluation of 14 representative image editing models on InEdit-Bench reveals significant and widespread shortcomings in this domain. By providing a standardized and challenging benchmark, we aim for InEdit-Bench to catalyze research and steer development towards more dynamic, reason-aware, and intelligent multimodal generative models.

\end{abstract}    
\vspace{-15pt}
\section{Introduction}
\label{sec:intro}
\vspace{-6pt}
Navigating a complex task is not a single, straightforward jump from inception to conclusion. Instead, the path to the solution comprises a series of indispensable intermediate steps that bridge the chasm between start and end~\cite{Zhou2022LeasttoMostPE}. Often, the true challenge lies not in the crossing itself, but in the fact that these ``key nodes'' are not readily apparent. We can perceive the starting point and final destination, yet the pathway connecting them remains invisible. Therefore, the ability to reconstruct this hidden path is a fundamental test of reasoning~\cite{Wei2022ChainOT,ZHENG2025TheCO}, prevalent across countless domains. 

This fundamental challenge of reasoning about transformative pathways is exemplified in intelligent image editing, which requires a shift from static outcomes~\cite{fang2025got,refBagel} (such as image creation or single-step edits) to modeling dynamic, procedural processes. While generative models~\cite{Vaswani2017AttentionIA,Ho2020DenoisingDP,Luo2023LatentCM,Esser2024ScalingRF,sdxl} demonstrate remarkable prowess in these static-oriented operations, this focus has consequently left their capacity for procedural reasoning largely underdeveloped. Even when exhibiting sophisticated semantic understanding~\cite{refEmu3,refAnyEditMU} or precise manipulation~\cite{Alaluf2021HyperStyleSI,refInstructPix2PixLT,xie2024showo,xie2025showo2}, this proficiency is typically confined to a single adjustment~\cite{refp2p,refomnigen2,refEmu2}, leaving their capacity for multi-step, sequential reasoning largely unaddressed. This raises a fundamental question from a modeling perspective: \textit{given only the initial and final images, how can a model generate an intermediate sequence that adheres to both causal consistency and visual plausibility?}

To bring greater attention to this unexplored frontier, we introduce a novel benchmark, \textbf{InEdit-Bench}, centered on the generation of intermediate logical pathways. 
Moving beyond conventional evaluations that simply appraise the final output~\cite{Sheynin2023EmuEP,Sushko2025RealEditRE,step1x-edit,Niu2025WISEAW}, our paradigm (see Fig.~\ref{Instruction_motivation}) requires a model to construct a sequence of transformations from an initial state to a final target.
This represents a significant departure from existing benchmarks~\cite{refRISE,refICE,refKRISE}, which, while valuable for assessing static outcomes such as instruction-following fidelity and semantic consistency~\cite{refICE,4v}, lack a quantitative measure of procedural reasoning. 
By shifting the evaluation focus from the ``destination'' to the ``intermediate logical pathways'', InEdit-Bench provides a more rigorous assessment of the core reasoning faculties, such as causal understanding and strategic planning. We aim to pivot the research focus from static, single-step outcomes towards the development of models capable of sophisticated procedural and dynamic reasoning.

Our InEdit-Bench consists of 237 high-quality, meticulously hand-annotated data instances. The dataset is organized into four fundamental categories: \textbf{{state transition}}, \textbf{{dynamic process}}, \textbf{{temporal sequence}}, and \textbf{{scientific simulation}}, collectively covering 16 distinct sub-tasks. Each benchmark instance comprises initial and final images with a textual prompt. To ensure a structured output, models are instructed to generate a single image of $N$ grids, where $N$ is adaptively determined to match the distinct stages of the process evolution. Additionally, every prompt includes a basic editing instruction and a brief summary of key intermediate steps, produced by an large multimodal model (LMM)~\cite{gpt-4o-2024-11-20} and human-checked.


For a comprehensive evaluation, InEdit-Bench employs six key metrics to assess the generated procedural pathways: \textbf{{appearance consistency}}, \textbf{{perceptual quality}}, \textbf{{semantic consistency}}, \textbf{{logical coherence}}, \textbf{{scientific plausibility}}, and \textbf{{process plausibility}}. While the first three metrics are adapted from standard image editing tasks~\cite{refICE,refRISE}, the latter three are novel and specifically designed for our process-oriented benchmark. These new metrics provide an objective assessment of the transition logic between stages, their scientific fidelity, and the holistic comprehension of the intermediate pathway. To automate this multifaceted evaluation, we adopt the {LMM-as-a-Judge} paradigm~\cite{refRISE,Niu2025WISEAW,wordgen}, where a powerful LMM serves as an objective evaluator for the generated image pathways.

Using InEdit-Bench, we conduct an assessment of 14 representative image editing models. The results uniformly demonstrate that these models exhibit significant shortcomings in multi-step editing and dynamic reasoning, highlighting a crucial direction for future research.


In summary, our main contributions are as follows:

(1) We introduce InEdit-Bench, the first benchmark for multi-step image editing and dynamic reasoning. It provides a challenging testbed to assess the ability of a model to comprehend and generate intermediate logical pathways, catalyzing future research in controllable visual editing.

(2)We constructed a meticulously annotated test set covering four fundamental task categories and established a six-dimensional evaluation protocol, providing a solid foundation for the systematic and rigorous assessment of model performance across complex editing paths.



(3) We present a comprehensive analysis of 14 state-of-the-art models on InEdit-Bench. Our findings reveal the significant limitations of current methods and highlight key areas for future improvement.

\begin{figure*}[t]
\centering
\includegraphics[width=0.985\textwidth]{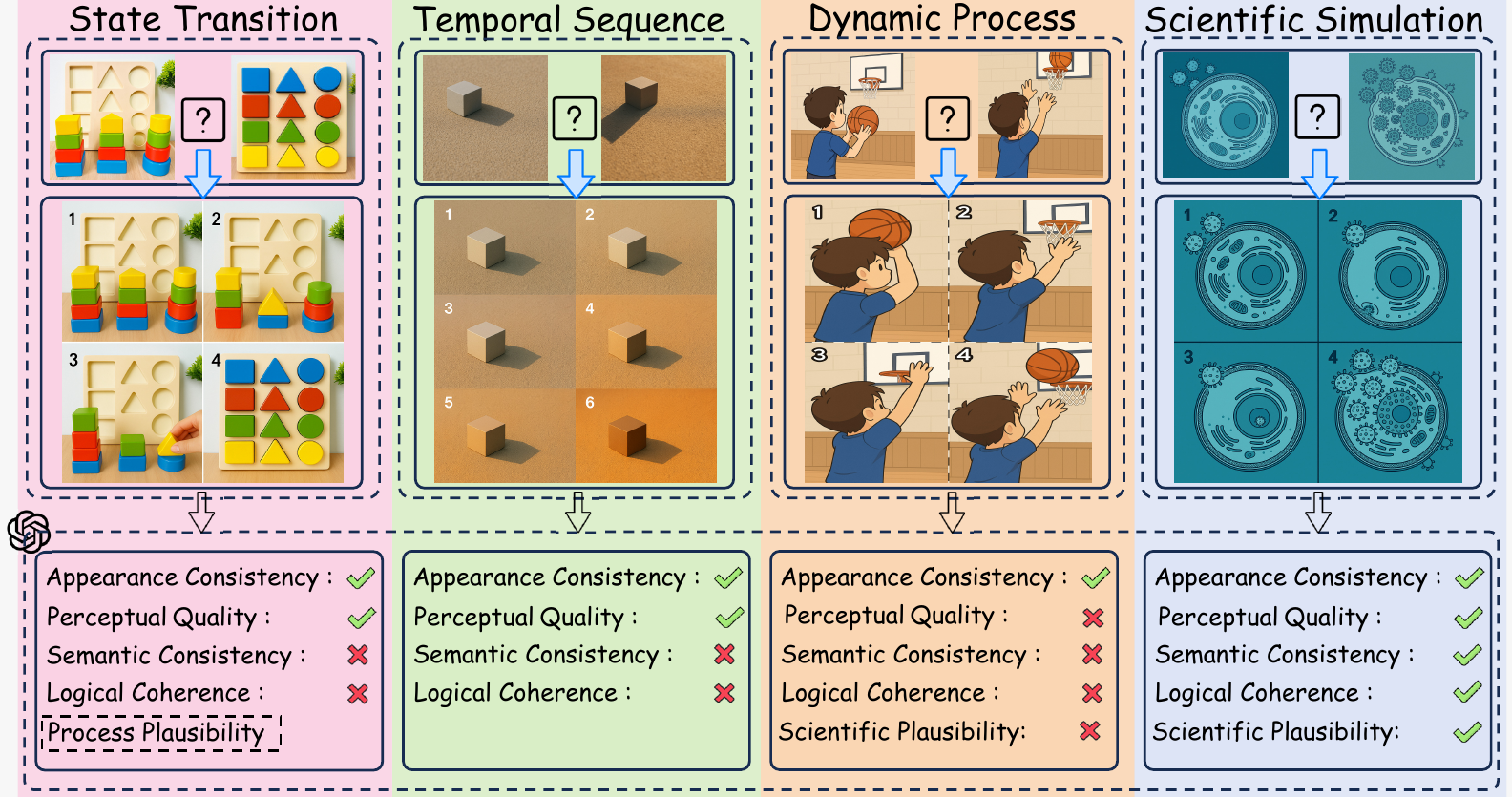} 
\caption{\textbf{Overall introduction to InEdit-Bench.} InEdit-Bench focuses on dynamic reasoning and multi-step editing modes, requiring models to generate intermediate logical pathways for given tasks. It spans 4 key domains: state transition, dynamic process, temporal sequence, and scientific simulation. The evaluation is conducted through 6 dimensions: appearance consistency, perceptual quality, semantic consistency, logical coherence, scientific plausibility, and process plausibility.}
\label{figIntroduction}
\vspace{-12pt}
\end{figure*}
\section{Related Work}
\label{Related Work}

\subsection{Instruction-Based Image Editing}
Instruction-based image editing (IIE)~\cite{refEmu3,ref1,instruct-edit2,Shuai2024ASO,Xu2025PersonalizedGI} simplifies user interaction~\cite{Li2023InstructAny2PixFV,Wang2023InstructEditIA} while improving controllability~\cite{refp2p,Avrahami2021BlendedDF,Hu2025PersonaHOIEI} and practicality~\cite{Zhang2023HIVEHH,He2024FreeEditMR}. Early work, such as InstructPix2Pix~\cite{refInstructPix2PixLT}, proposed driving image editing with simple instructions, and subsequent research has continuously improved data quality and model architectures. MagicBrush~\cite{magicbrush} introduced high-quality manually annotated data, MGIE~\cite{refMGIE} and SmartEdit~\cite{refSmartEdit} leveraged multimodal large language models to enhance semantic understanding and reasoning, while OmniGen~\cite{refomnigen2} and Gemini~\cite{geminiteam2025} built unified multimodal architectures to strengthen task generalization. While existing methods have achieved progress in quality~\cite{sdxl,highr}, efficiency~\cite{mixed,janus}, and controllability~\cite{tune}, their capabilities for multi-step editing and higher-order understanding remain largely underexplored. To address this, we propose InEdit-Bench.

\subsection{Image Editing Benchmarks}
Recent benchmarks like TedBench~\cite{tedbench}, EditVal~\cite{EditVal}, and EditBench~\cite{editbench} primarily focus on basic editing tasks, while MagicBrush~\cite{magicbrush} provides high-quality data but its evaluation metrics~\cite{DINO,CLIP} have inherent limitations in reflecting image quality. With further research, Reason-edit~\cite{refSmartEdit} and RISEBench~\cite{refRISE} emphasize evaluating models’ understanding and reasoning capabilities under complex instructions. Complex-Edit~\cite{Complex-Edit} introduces a chain-of-thought-like multi-step editing mechanism, while I2Ebench~\cite{I2EBench} and KRIS-Bench~\cite{refKRISE} explore higher-level editing abilities from the perspectives of multi-dimensional skills and knowledge-driven reasoning. Overall, despite advances in scale~\cite{Li2024SEEDBenchBM, refICE} and diversity~\cite{refRISE,refKRISE}, existing benchmarks remain insufficient in modeling intermediate logical pathways and supporting dynamic chain-style reasoning.~\cite{Sheynin2023EmuEP,Sushko2025RealEditRE,step1x-edit}. To address this gap, we propose InEdit-Bench, designed to systematically evaluate intelligent visual editing models in multi-step image editing and dynamic reasoning.

\begin{table}[h]
\centering
\footnotesize
\setlength{\tabcolsep}{2.6pt}
\renewcommand\arraystretch{1.1} 
\vspace{-3pt}
\caption{Comparison of open-source reasoning-based image editing benchmarks. T-C-S-L represents Time-Cause-Space-Logic.}
\vspace{-2pt}
\label{comparison_bench}
\begin{tabular}{lcccc}
\toprule
\textbf{Dataset} & \textbf{Publication} & \textbf{Size} & \textbf{Dimensions} & \textbf{Perspective}  \\
\midrule
AURORA-Bench~\cite{aurora-Bench}        & NeurIPS 2024 & 400 & --   & Action  \\
SmartEdit~\cite{refSmartEdit}       & CVPR 2024   & 219 & 2    & Attribute  \\
RISEBench~\cite{refRISE}                  & NeurIPS 2025 & 360 & 4   & T-C-S-L  \\
IntelligentBench~\cite{refBagel} & arXiv 2025.5 & 350 & --  & Deep  \\
\textbf{InEdit-Bench}               & --          & 237 & 4  & Pathway \\
\bottomrule
\end{tabular}
\end{table}

\vspace{-6pt}

\section{InEdit-Bench}
\subsection{Benchmark Construction}

\begin{figure}[t]
  \centering
  \includegraphics[width=0.41\textwidth]{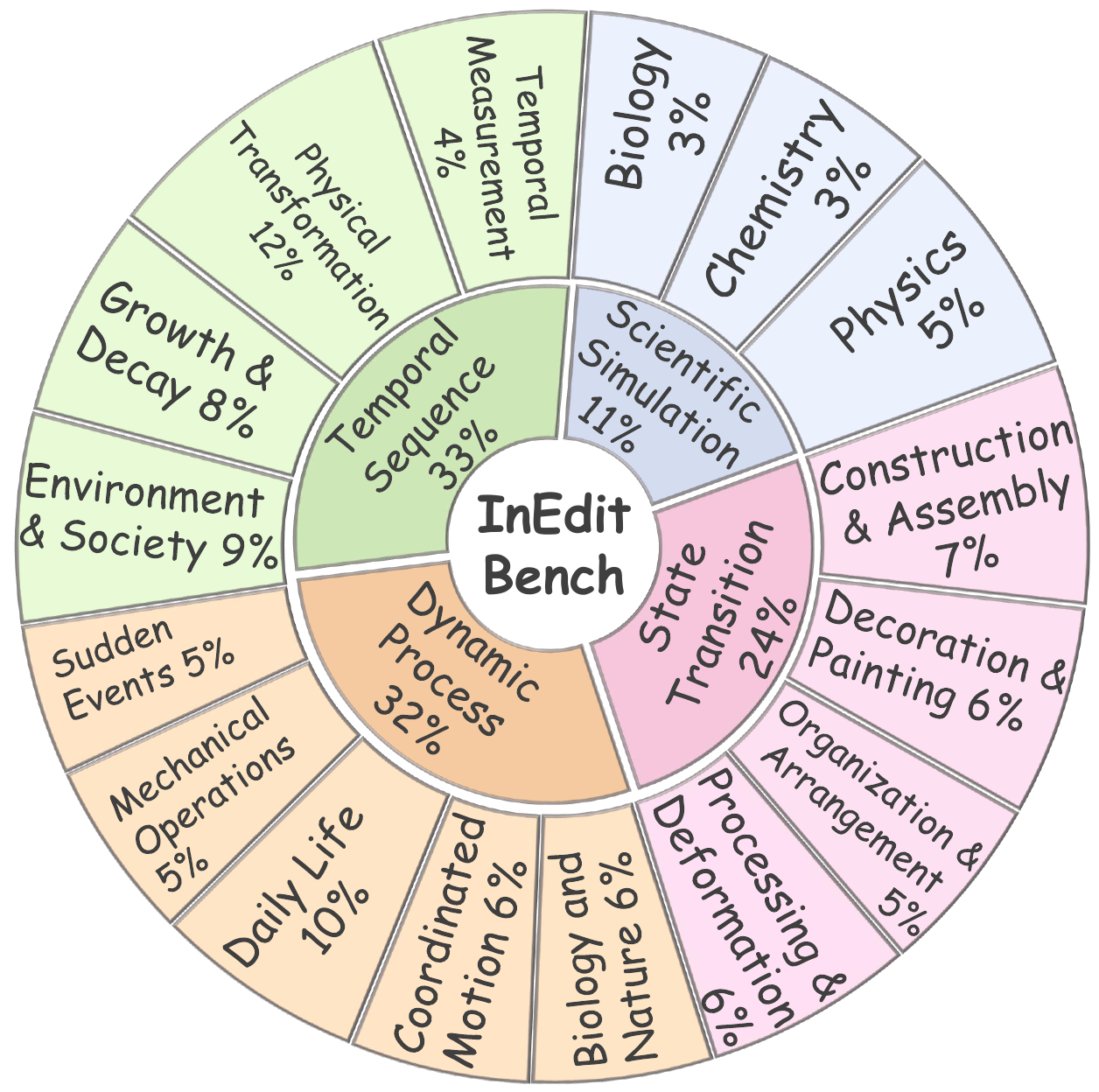}

   \caption{The task type distribution of InEdit-Bench. InEdit-Bench conducts a comprehensive evaluation of visual editing models across 16 sub-tasks under 4 domains.}
   \label{figbench}
   \vspace{-12pt}
\end{figure}

Tab.~\ref{comparison_bench} presents a comparative analysis with previous reasoning-based image editing benchmarks. InEdit-Bench is an innovative benchmark designed to systematically evaluate a model's capability in comprehending and representing intermediate logical pathways. As illustrated in Fig.~\ref{figIntroduction}, we categorize editing tasks into four types based on the evolutionary dynamics of their intermediate states: \textbf{{state transition}}, \textbf{{dynamic process}}, \textbf{{temporal sequence}}, and \textbf{{scientific simulation}}. Fig.~\ref{figbench} provides a detailed illustration of the task distribution within InEdit-Bench.
\noindent\textbf{State Transition.}
State transition reasoning infers and reconstructs discrete changes between initial and final states. The primary challenge is identifying key nodes and dependencies to generate a logically coherent editing sequence. This task encompasses four subcategories. \textbf{(1)~Construction and Assembly:} This subtask requires the model to combine independent components into a complete entity based on spatial logic. The challenge lies in structural reconstruction and resolving assembly dependencies. \textbf{(2)~Decoration and Painting:} This involves applying colors and patterns to a surface. The key lies in precise region identification and operation sequencing. \textbf{(3)~Organization and Layout:} The focus is on evaluating the model’s ability to systematically organize elements and manage spatial layout. \textbf{(4)~Processing and Deformation:} Involves altering object morphology or structure through discrete steps. The challenge lies in precisely simulating the discontinuous yet logically coherent topological evolution (\textit{e.g.}, origami).

\vspace{4pt}

\noindent\textbf{Dynamic Process.}
Dynamic process reasoning involves continuous transformations from initial to final states. Unlike discrete state transitions, it requires handling fluid progressions where each intermediate step maintains natural fluidity and logical consistency. This category includes five subdomains. \textbf{(1)~Biology and Nature:} Focuses on organism evolution and natural phenomena, requiring deduction from biological and natural laws (\textit{e.g.}, a spider weaving a web). \textbf{(2)~Coordinated Motion:} Involves fluid entity movement through space, requiring the model to comprehend motion dynamics for smooth, logically connected actions (\textit{e.g.}, a long jump). \textbf{(3)~Daily Life:} Encompasses modeling common interactions and behaviors in everyday contexts. \textbf{(4)~Mechanical Operation:} Illustrates continuous structural alterations via incremental reasoning, highlighting the driving operational mechanisms (\textit{e.g.}, a compressor flattening a cube). \textbf{(5)~Sudden Events:} Features abrupt, often destructive transformations, requiring the identification of pivotal moments to render believable visual consequences (\textit{e.g.}, a building demolition).

\begin{figure*}[!ht]
\centering
\includegraphics[width=0.98\textwidth]{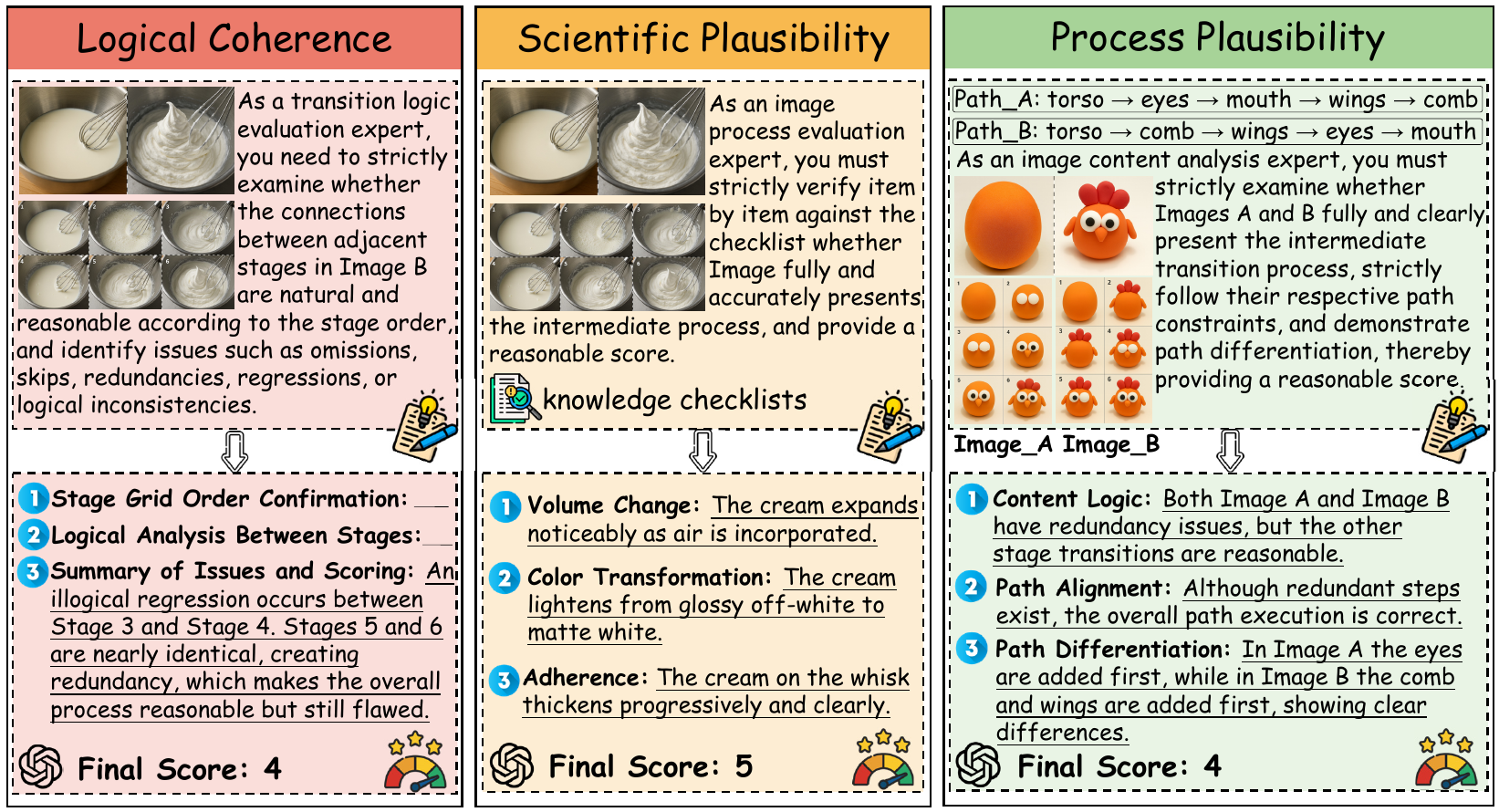} 
\caption{The evaluation metrics of \textbf{Logical Coherence}, \textbf{Scientific Plausibility}, and \textbf{Process Plausibility} in InEdit-Bench. For each evaluation dimension, the evaluator model (GPT-4o-2024-11-20~\cite{gpt-4o-2024-11-20} in this study) analyzes various inputs based on carefully designed prompts and assigns corresponding scores for each sub-dimension.}
\vspace{-12pt}
\label{evalution_metric}
\end{figure*}

\vspace{4pt}

\noindent\textbf{Temporal Sequence.}
Temporal sequence reasoning focuses on the gradual evolution of target states over time. Unlike dynamic process, which emphasize mechanism-driven continuous actions (how it happens), temporal tasks focus on timeline-driven phase evolution (when or along which phases it happens). This domain is subdivided into four subcategories. \textbf{(1)~Environment and Society:} Concerns the gradual evolution of environmental and social phenomena (\textit{e.g.}, sand dune formation). \textbf{(2)~Growth and Decay:} Pertains to biological life cycles (\textit{e.g.}, a flower blooming), requiring time-series generation aligned with biological principles. \textbf{(3)~Physical Transformation:} Involves changes in material or object properties over time (\textit{e.g.}, melting glaciers). The primary challenge lies in modeling their temporal progression. \textbf{(4)~Temporal Measurement:} Focuses on time as a quantifiable metric (\textit{e.g.}, an hourglass, a progress bar), demanding precise reasoning about quantitative temporal changes.

\vspace{4pt}

\noindent\textbf{Scientific Simulation.}
Scientific simulation is designed to model principles from the fields of \textbf{Physics}, \textbf{Chemistry}, and \textbf{Biology}, strictly requiring adherence to scientific laws while illuminating intermediate logical steps. Examples encompass physical phenomena (\textit{e.g.}, diffusion, total internal reflection), chemical reactions (\textit{e.g.}, magnesium combustion, displacement), and biological processes (\textit{e.g.}, cell division, DNA replication). These tasks require the model to comprehend complex scientific procedures, deduce causal mechanisms, and accurately render pivotal stages. To streamline this process, task instructions provide concise keyframe prompts, ensuring the model concentrates on the most critical phases and disregards superfluous steps.

\subsection{Evaluation Metrics}

Diverging from conventional benchmarks that assess single-step image editing, InEdit-Bench shifts the evaluation focus from input-output comparisons to the procedural integrity of the entire transformation process. To this end, we develop a multi-faceted evaluation framework built upon six key dimensions. These are categorized into two groups: three foundational metrics for visual quality (\textbf{Appearance Consistency}, \textbf{Perceptual Quality}, \textbf{Semantic Consistency}), and three novel dimensions (see Fig.~\ref{evalution_metric}) designed to scrutinize the plausibility of intermediate processes (\textbf{Logical Coherence}, \textbf{Scientific Plausibility}, \textbf{Process Plausibility}). Our evaluation employs the LMM-as-a-Judge methodology, utilizing GPT-4o as the evaluator to enable automated assessment. During the evaluation process, the evaluator receives the user instructions, scoring rubric, and the generated output, based on which it provides a numerical score for each dimension.

\begin{table*}[t]
  \centering
  \caption{
  \textbf{Performance of various models on InEdit-Bench.} The best and second-best performances of the proprietary and open-source models are highlighted in \textbf{bold} and with an \underline{underline}, respectively. 95\% confidence intervals (CIs) are estimated via stratified bootstrapping (10k iterations). Accuracy denotes the percentage of ``perfect'' samples where all evaluation metrics reach the maximum score.
  }
  \label{table1}
  \small
  \setlength{\tabcolsep}{4pt}
  \renewcommand\arraystretch{1.0} 
  \setlength{\tabcolsep}{1.75pt}
  \begin{tabular}{l|cccccc|ccc}
    \toprule
    \textbf{Models} & \makecell{\textbf{Appearance}\\\textbf{Consistency}} & \makecell{\textbf{Perceptual}\\\textbf{Quality}} & \makecell{\textbf{Semantic}\\\textbf{Consistency}} & \makecell{\textbf{Logical}\\\textbf{Coherence}} & \makecell{\textbf{Scientific}\\\textbf{Plausibility}} & \makecell{\textbf{Process}\\\textbf{Plausibility}} & 
    \makecell{\textbf{Overall}\\\textbf{Average}} &
    \makecell{\textbf{Bootstrap }\\\textbf{95\% CIs}} &
    \textbf{Accuracy}\\
    \midrule
    \multicolumn{10}{c}{\textbf{Proprietary Models}} \\
    \midrule
    GPT-Image-1 & \textbf{92.24} &	\underline{92.36} &	\textbf{72.04} &	\textbf{71.06} &	\underline{71.31} &	\textbf{88.97} &	\textbf{81.33} & \textbf{[79.04, 83.61]} & \textbf{16.75\%} 
 \\
    Nano-Banana & \underline{86.45} &	\textbf{92.49} &	\underline{62.93} &	\underline{60.22} &	\textbf{73.58} &	\underline{75.74}   & \underline{75.23} & \underline{[72.40, 77.96]} & \underline{13.30\%}
\\
    Flux-Kontext-pro & 64.66 &	89.11 &	33.99 &	30.17 &	43.75 &	47.06 &	51.46 & [48.59, 54.45] &	0.99\% 
 \\
    Doubao-SeedEdit-3.0 & 44.43 &	69.70 &	22.54 &	22.41 &	34.94 &	25.00  & 36.50 & [34.04, 39.10] &	0.00\% 
\\
    \midrule
    \multicolumn{10}{c}{\textbf{Open-Source Models}} \\
    \midrule
    Qwen-Image-Edit & \textbf{62.32} &	\underline{82.64} &	\underline{27.34} &	\textbf{28.94} &	\textbf{44.89} &	\textbf{51.47} &	\textbf{49.60} & \textbf{[46.87, 52.43]} &	\underline{0.49\%} 
 \\
    Emu1 & 5.17 &	48.65 &	2.46 &	3.45 &	5.11 &	3.68 &	11.42 & [10.36, 12.57] &	0.00\% 
 \\
    Emu2 & 33.17 &	\textbf{85.30} &	6.16 &	15.15 &	22.44 &	15.44  &	29.61 & [27.61, 31.81] &	0.00\% 
\\
    Bagel & 46.18 &	65.89 &	\textbf{28.08} &	\underline{27.34} &	\underline{34.09} &	\underline{42.65} &	\underline{40.70} & \underline{[37.99, 43.49]} &	0.00\% 
\\
    Bagel-Think & \underline{53.94} &	76.72 &	24.01 &	\textbf{28.94} &	\underline{34.09} &	26.47 &	\underline{40.70} & \underline{[37.99, 43.54]} &	\textbf{0.99\%} 
 \\
    OmniGen & 9.24 &	35.71 &	5.42 &	7.76 &	13.92 &	13.97  &	14.34 & [12.55, 16.29] &	0.00\% 
\\
    OmniGen2 & 42.36 &	78.94 &	21.31 &	24.75 &	29.26 &	30.88 &	37.92 & [35.16, 40.78] &	\underline{0.49\%} 
\\
    Step1X-Edit(v1.0) & 15.89 &	42.66 &	7.39 &	8.00 &	15.06 &	9.56 
  &	16.43 & [14.48, 18.53] &	0.00\% 
\\
    Step1X-Edit(v1.1) & 34.61 &	54.56 &	17.00 &	23.89 &	31.82 &	26.47 
 &	31.39 & [28.72, 34.18] &	0.00\% 
\\
    InstructPix2Pix & 33.62 &	74.50 &	4.46 &	13.42 &	13.35 &	0.00 
 &	23.23 & [21.75, 24.68] &	0.00\% 
\\
    \bottomrule
  \end{tabular}
  \vspace{-8pt}
\end{table*}

\subsection{Standard Visual Quality Metrics}
To establish a baseline for visual quality, our framework incorporates three foundational metrics from the image editing domain. Appearance consistency assesses the preservation of style and visual attributes across all depicted stages of the process. Perceptual quality measures the realism and fidelity of the generated imagery, ensuring it is free from artifacts. Semantic consistency evaluates the alignment of the final image content with the specified editing objective. Collectively, these metrics provide a robust assessment of the visual integrity of the final output.


\subsection{Proposed Process-Oriented Metrics}

\textbf{Logical Coherence.}
Logical coherence is paramount in evaluating multi-step image editing, as it examines the integrity of the generated process in terms of its logical progression and natural flow. The assessment protocol begins by establishing the sequence of the depicted stages, applying a top-to-bottom, left-to-right convention whenever the intended order is not visually apparent. The core of the evaluation then involves a close examination of the transitions between adjacent stages for logical soundness and naturalness. This scrutiny ensures that the overall evolution is fluid and coherent, devoid of any jarring discontinuities or superfluous, repetitive actions.

\vspace{4pt}
\noindent\textbf{Scientific Plausibility.}
Drawing inspiration from KRIS-Bench~\cite{refKRISE} and WorldGenBench~\cite{wordgen}, we incorporate scientific plausibility as a dedicated evaluation dimension. This metric is applied to tasks involving dynamic process and scientific simulation, where adherence to scientific logic is assessed using manually designed knowledge checklists. These checklists annotate the key features or inherent mechanisms that should be present in the intermediate stages. The evaluation process involves comparing the generated visual content against the items in the checklist to verify its compliance with predefined scientific standards.

\vspace{4pt}
\noindent\textbf{Process Plausibility.}
To comprehensively evaluate how well a model understands intermediate pathways, we employ two prompting schemes with distinct path constraints, applied to a subset of our state transition tasks. This approach is inspired by the fact that many real-world processes are non-deterministic, often presenting multiple viable paths to the same outcome. Therefore, a robust model should be able to understand multiple feasible paths for the same process while ensuring consistency and accuracy.


Therefore, we introduce process plausibility as an advanced metric. Under this metric, successful generation must adhere to the global path sequence, avoiding logical errors or expression inconsistencies between stages, thereby ensuring a clear depiction of the intended logical trajectory. By observing the model's performance under the two constraint schemes, we analyze the model's overall grasp of path constraints and its ability to follow the specified path.

\begin{figure*}[t]
\centering
\includegraphics[width=0.98\textwidth]{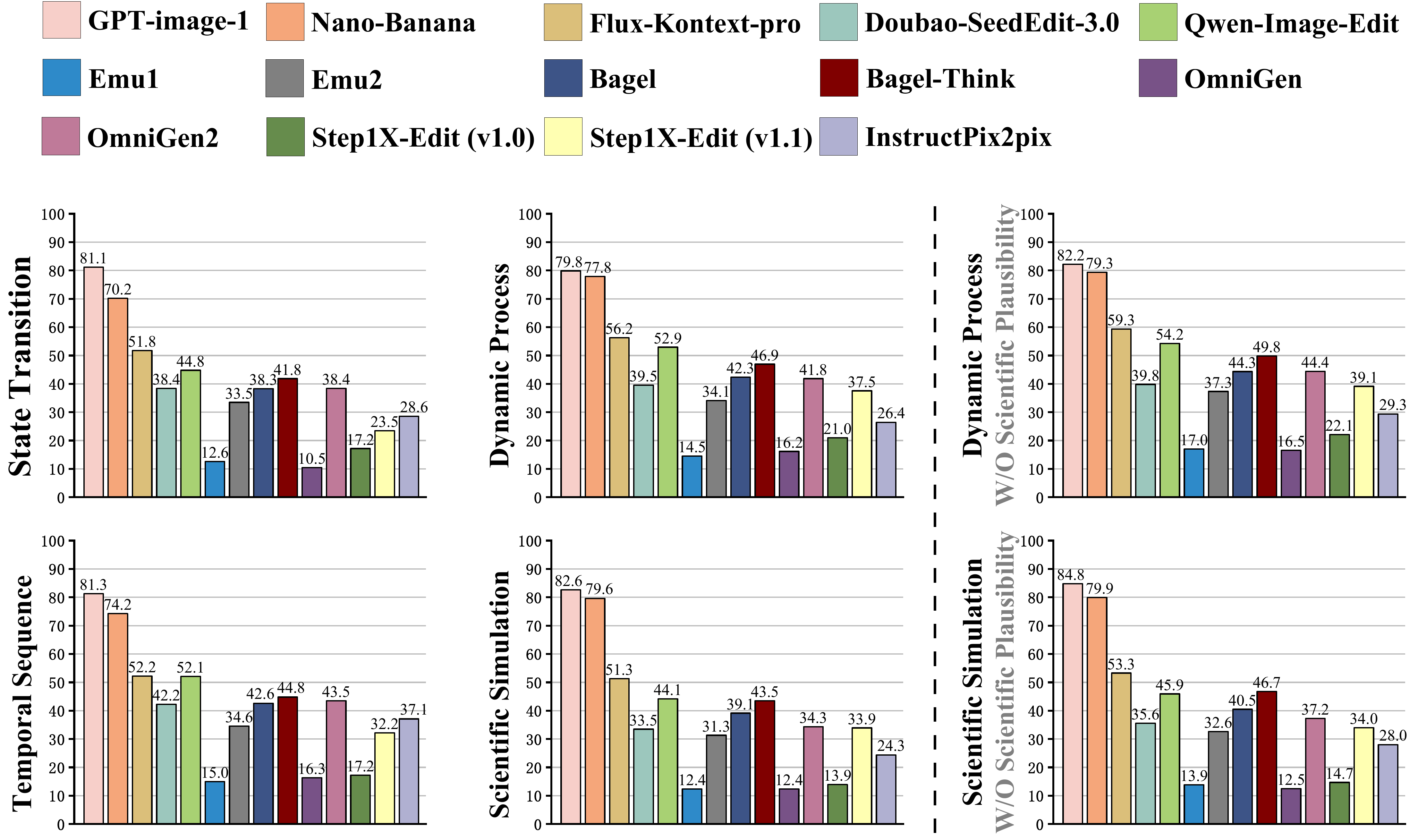} 
\caption{Comparison of models across four fundamental tasks. For the dynamic process and scientific simulation tasks, scores in gray denote performance calculated without the \textit{scientific plausibility} metric.}
\label{four-base-tasks-1013}
\vspace{-10pt} 
\end{figure*}
\section{Experiments}
\subsection{Experiments Setup}
On InEdit-Bench, we evaluated 14 representative visual editing models. The evaluated models include proprietary models: GPT-Image-1~\cite{gpt-image-1}, Nano-Banana~\cite{nano-banana}, Flux-Kontext-pro~\cite{flux-kongtext-pro}, and Doubao-SeedEdit-3.0-i2i~\cite{doubao}; as well as open-source models: Qwen-Image-Edit~\cite{Qwen-Image}, Bagel~\cite{refBagel}, OmniGen~\cite{omnigen1}, OmniGen2~\cite{refomnigen2}, Step1X-Edit~\cite{step1x-edit}, Emu1~\cite{emu1}, Emu2~\cite{refEmu2}, and InstructPix2Pix~\cite{refInstructPix2PixLT}. These models cover a range of mainstream generative architectures, including autoregressive generation paradigms~\cite{refEmu2,refBagel}, diffusion model architectures~\cite{refInstructPix2PixLT}, and diffusion transformer architectures~\cite{step1x-edit}. All generation and evaluation processes are conducted on L20 GPUs using default hyperparameter settings. Additionally, since some open-source models do not support multi-image input, we uniformly concatenate the initial image and the final image into a single image as input for all models.

\subsection{Result Analysis}
\noindent\textbf{Results Analysis by Metrics.}
Tab.~\ref{table1} reports the scores of 14 models on InEdit-Bench. All scores are normalized to a 100-point scale and are evaluated by GPT-4o-2024-11-20~\cite{gpt-4o-2024-11-20}. Furthermore, the narrow 95\% CIs derived from stratified bootstrapping confirm the statistical stability of the reported scores. Results on InEdit-Bench show that GPT-Image-1 is the best-performing proprietary model, with a score of 81.33 and an accuracy of 16.75\%. Among open-source models, Qwen-Image-Edit and Bagel-Think perform relatively well, scoring 49.60 and 40.70 points, respectively. Although there is still a gap between open-source and proprietary models, some open-source models have nevertheless demonstrated outstanding capabilities.

The experiments indicates that proprietary models, specifically GPT-Image-1 and Nano-Banana, consistently lead in appearance consistency, semantic consistency, and logical coherence, demonstrating their robust and well-rounded capabilities. While open-source models post lower aggregate scores, several demonstrate significant potential in specific dimensions. For instance, Qwen-Image-Edit stands out among open-source solutions for the high performance in semantic consistency, logical coherence, and scientific plausibility, occasionally rivaling proprietary counterparts. Similarly, the Bagel series is highly competitive in perceptual quality and semantic consistency, a strength also exhibited by OmniGen2 in perceptual quality.

On the other hand, open-source models Emu1 and OmniGen struggle to maintain effective visual consistency. Among proprietary models, Doubao lags behind, with significantly lower scores in appearance consistency, semantic consistency, and logical coherence compared with its peers. This suggests that Doubao may place more emphasis on rapid local editing while lacking robustness in modeling global consistency and cross-modal logical constraints. Notably, half of the open-source models score below 10.00 in the semantic consistency dimension, further highlighting their systematic deficiencies in intermediate logical path reasoning and editing.

In terms of process plausibility, GPT-Image-1 achieves the highest score of 89.00, demonstrating its advantage in understanding and articulating reasoning paths, with Nano-Banana ranking closely behind. In contrast, most of the other models generally struggle to fully meet task requirements. With respect to accuracy, even the best-performing model, GPT-Image-1, achieves only 16.75\%, with Nano-Banana closely following at 13.30\%. The remaining models have accuracy rates below 1.00\%, with several models scoring 0\%. Overall, these results reveal that current models still face significant limitations in long-term dependency capture and multi-stage causal reasoning. Achieving accurate modeling and representation of intermediate reasoning paths remains a key challenge that urgently needs to be addressed in the future.

\begin{figure*}[t!]
\centering
\includegraphics[width=0.978\textwidth]{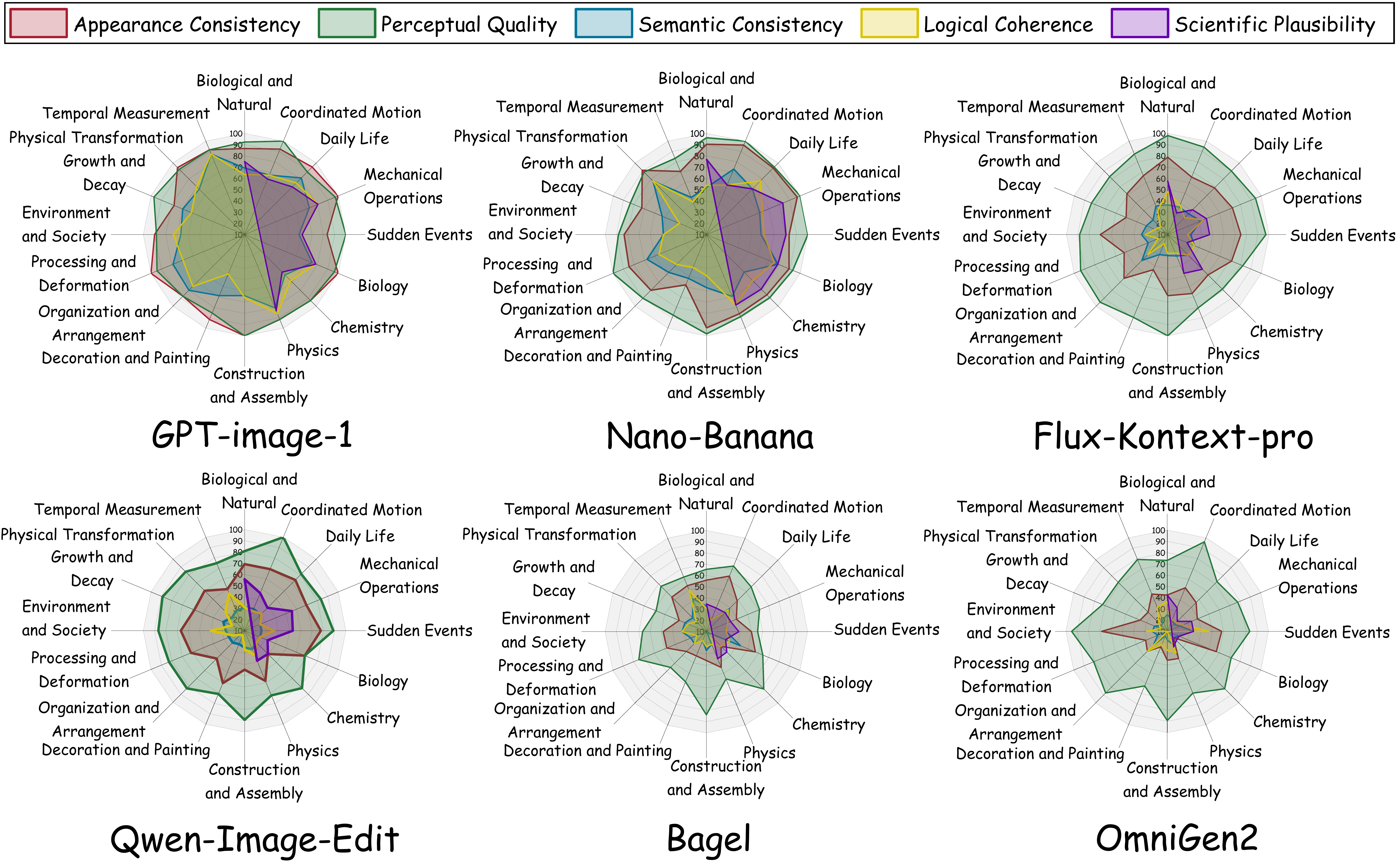} 
\caption{Performance of GPT-image-1, Nano-Banana, Flux-Kontext-pro, Qwen-Image-Edit, Bagel and OmniGen2 across 16 sub-tasks.}
\label{figradar}
\vspace{-5pt} 
\end{figure*}

\vspace{4pt}
\noindent\textbf{Results Analysis by Tasks.}
From the perspective of task dimensions, state transition task poses significant challenges for models. As shown in Fig.~\ref{four-base-tasks-1013}, the scores of all models in this category are lower than their performance on temporal sequence and dynamic process tasks. Notably, nine models, including GPT-Image-1 and Nano-Banana, score lower on state transition task than on scientific simulation task (when comparing state transition, dynamic process, and scientific simulation tasks, only the average scores of four metrics—appearance consistency, perceptual quality, semantic consistency, and logical coherence—are uniformly calculated). Furthermore, apart from GPT-Image-1 and Nano-Banana, all other models achieve lower average scores on scientific simulation task compared to their performance on dynamic process task. These phenomena reveal a layered structure of task complexity: from continuous to discrete, and from surface-level phenomena to deeper scientific principles, model performance shows a step-by-step decline. This highlights the limitations of current large models in complex logical reasoning and scientific law modeling.

Fig.~\ref{figradar} presents the performance of several representative models on 16 subtasks. Overall, GPT-Image-1 achieved stable and superior performance. In contrast, the results of Flux-Kontext-pro, Qwen-Image-Edit, Bagel, and OmniGen2 exhibit significant fluctuations, particularly in the dimensions of semantic consistency and logical coherence, where substantial performance gaps may arise even among subtasks within the same basic task category. Specifically, within the temporal sequence category, Flux-Kontext-pro and Bagel experience a notable drop in performance on the physical transformation subtask, while Qwen-Image-Edit and OmniGen2 show marked degradation on the growth and decay subtask. Nano-Banana achieves relatively better average performance on dynamic process and scientific simulation tasks. However, it still suffers from performance decline and uneven results in the state transition and temporal sequence categories.

\subsection{Validity of LMM Scores}
\begin{figure}[ht]
  \centering
  \includegraphics[width=0.5\textwidth]{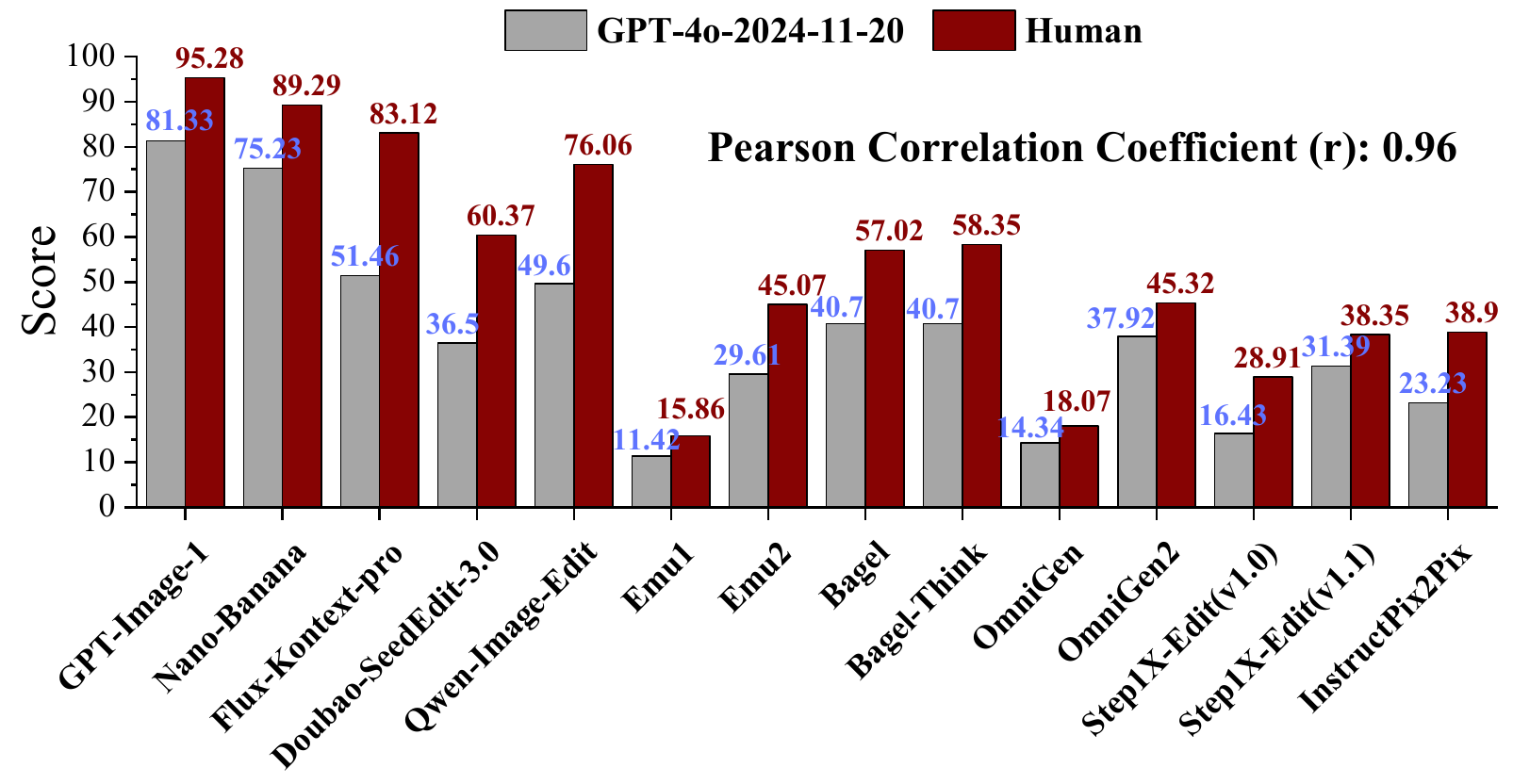}

   \caption{Comparison of LMM automated and human evaluation results, with their Pearson correlation coefficient ($r$).}
   \label{humanandgpt}
   \vspace{-5pt}
\end{figure}

To evaluate LMM score reliability, we collected 168 evaluation records from 12 annotators. Each record consists of an annotator’s ranking of 14 model outputs for a single test sample. We applied the Borda Count method to analyze these assessments, with the final scores normalized to a 100-point scale. Fig.~\ref{humanandgpt} compares the automated evaluation results of the proposed LMM with the human evaluation results, presenting the Pearson correlation coefficient ($r$). The results show a high correlation ($r = 0.96$) between the automated and human evaluations, validating the reliability of our proposed LMM-based automated evaluation method.

\section{Conclusion}
In this paper, we propose InEdit-Bench, the first system evaluation benchmark focused on image multi-step editing and intermediate logical path reasoning. It covers four basic categories: state transition, dynamic process, temporal sequence, and scientific simulation. Based on this, we design six evaluation dimensions: appearance consistency, perceptual quality, semantic consistency, logical coherence, scientific plausibility, and process plausibility, to comprehensively measure the ability of intelligent visual editing models in intermediate logical path reasoning and expression. Through the evaluation of 14 representative models, we reveal significant shortcomings in current models regarding multi-step editing and dynamic reasoning capabilities, providing clear directions and reference for further optimization of model performance.
\clearpage
{
    \small
    \bibliographystyle{ieeenat_fullname}
    \bibliography{main}
}

\appendix
\clearpage
\setcounter{page}{1}
\maketitlesupplementary

%
%
%

\section{Overview of Supplementary Material}
This supplementary material supplements the proposed InEdit-Bench with details excluded from the main paper due to space constraints.

The supplementary material is organized as follows:
\begin{itemize}
    \item Sec. ~\ref{More Detailed Evaluation Results}: More Detailed Evaluation Results.
    \item Sec. ~\ref{Human Evaluator}: Human Evaluator.
    \item Sec. ~\ref{Data Source of InEdit-Bench}: Data Source of InEdit-Bench.
    \item Sec. ~\ref{Limitations}: Limitations.
    \item Sec. ~\ref{Representative Example Images from InEdit-Bench}: Representative Example Images from InEdit-Bench.
    \item Sec. ~\ref{Detailed Outputs of Evaluated Models}: Detailed Outputs of Evaluated Models.
    \item Sec. ~\ref{Design of the Prompt}: Design of the Prompt.
\end{itemize}

\section{More Detailed Evaluation Results} \label{More Detailed Evaluation Results}
In this section, we present a more detailed evaluation of the models, offering further analysis of their capabilities. This includes:

(1) The specific scores of 14 models across 4 fundamental tasks and 16 sub-tasks.

(2) The accuracy of these 14 models across the 16 sub-tasks.

\subsection{Scores of Models Across 4 Tasks and 16 Sub-Tasks}
\setlength{\tabcolsep}{5pt} 
\begin{table*}[th!]
\renewcommand\arraystretch{1.15}
\caption{The specific scores of the models across four fundamental tasks, with metrics including Appearance Consistency (AC), Perceptual Quality (PQ), Semantic Consistency (SC), Logical Coherence (LC), Scientific Plausibility (SP). The performance of open-source and proprietary models is separately marked with the best performance in \textbf{bold}, and the second best \underline{underlined}.}
\label{Specific Scores Across Four Four Fundamental Tasks}
\centering
\resizebox{\linewidth}{!}{
\begin{tabular}{cccccccccccccccccc}
\specialrule{0.1em}{0pt}{2pt}
\multicolumn{2}{c}{\multirow{2}{*}[-7ex]{Metric}} & & \multicolumn{4}{c}{\textbf{Proprietary Models}} & & \multicolumn{10}{c}{\textbf{Open-Source Models}} \\ 
\cmidrule{4-7} \cmidrule{9-18}
& & & \multicolumn{1}{c}{\rotatebox{90}{\small GPT-Image-1}} &
\multicolumn{1}{c}{\rotatebox{90}{\small Nano-Banana}} &
\multicolumn{1}{c}{\rotatebox{90}{\small Flux-Kontext-pro}} &
\multicolumn{1}{c}{\rotatebox{90}{\small Doubao-SeedEdit-3.0}} & &
\multicolumn{1}{c}{\rotatebox{90}{\small Qwen-Image-Edit}} &
\multicolumn{1}{c}{\rotatebox{90}{\small Emu1}} &
\multicolumn{1}{c}{\rotatebox{90}{\small Emu2}} &
\multicolumn{1}{c}{\rotatebox{90}{\small Bagel}} &
\multicolumn{1}{c}{\rotatebox{90}{\small Bagel-Think}} &
\multicolumn{1}{c}{\rotatebox{90}{\small OmniGen}} &
\multicolumn{1}{c}{\rotatebox{90}{\small OmniGen2}} &
\multicolumn{1}{c}{\rotatebox{90}{\small Step1X-Edit (v1.0)}} &
\multicolumn{1}{c}{\rotatebox{90}{\small Step1X-Edit (v1.1)}} &
\multicolumn{1}{c}{\rotatebox{90}{\small InstructPix2Pix}} \\
\specialrule{0.07em}{0pt}{0pt}
\multirow{5}{*}{\makecell{\emph{\textbf{State}}\\ \emph{\textbf{Transition}}}}

& AC  &  & \textbf{95.41} &	\underline{79.59} &	56.12 &	42.71 & &	\textbf{53.06} &	3.06 &	30.10 &	38.27 &	\underline{48.47} &	5.61 &	31.63 &	12.76 &	21.94 &	24.49 
\\

& PQ &   & 92.35 &	\underline{94.39} &	\textbf{94.79} &	69.39 & &	\underline{81.63} &	44.90 &	\textbf{83.33} &	69.39 &	78.06 &	29.08 &	80.10 &	42.71 &	47.45 &	72.96 
 \\
& SC  &  & \textbf{72.45} &	\underline{58.67} &	31.63 &	21.94 & &	\textbf{24.49} &	1.53 &	8.67 &	\underline{23.98} &	19.90 &	2.04 &	19.39 &	7.65 &	10.71 &	4.59 
 \\
& LC  &  & \textbf{64.29} &	\underline{47.96} &	24.49 &	19.39 & &	19.90 &	1.02 &	11.73 &	\underline{21.43} &	20.92 &	5.10 &	\textbf{22.45} &	5.61 &	13.78 &	12.24 
 \\
& Avg &  & \textbf{81.12} &	\underline{70.15} &	51.76 &	38.36 & &	\textbf{44.77} &	12.63 &	33.46 &	38.27 &	\underline{41.84} &	10.46 &	38.39 &	17.18 &	23.47 &	28.57 
 \\
\specialrule{0.07em}{0pt}{0pt}

\multirow{5}{*}{\makecell{\emph{\textbf{Temporal}}\\ \emph{\textbf{Sequence}}}} 
& AC  &  & \textbf{89.39} &	\underline{85.23} &	64.39 &	49.24 & &	\textbf{62.12} &	6.82 &	35.00 &	47.73 &	\underline{55.68} &	9.47 &	47.73 &	16.29 &	35.23 &	45.08 
 \\
& PQ  &  & \textbf{89.77} &	\underline{87.12} &	83.71 &	69.32 & &	\underline{83.33} &	45.45 &	\textbf{87.30} &	64.39 &	75.00 &	40.15 &	79.55 &	37.50 &	50.76 &	78.52 
 \\
& SC  &  & \textbf{71.59} &	\underline{64.39} &	33.71 &	23.11 & &	\underline{28.41} &	2.65 &	1.52 &	\textbf{29.17} &	20.83 &	6.06 &	21.21 &	6.82 &	16.67 &	5.38 
 \\
& LC  &  & \textbf{74.24} &	\underline{60.23} &	26.89 &	27.27 & &	\textbf{34.47} &	4.92 &	14.39 &	\underline{29.17} &	27.65 &	9.47 &	25.38 &	8.33 &	26.14 &	19.32 
 \\
& Avg &  & \textbf{81.25} &	\underline{74.24} &	52.18 &	42.23 & &	\textbf{52.08} &	14.96 &	34.55 &	42.61 &	\underline{44.79} &	16.29 &	43.47 &	17.23 &	32.20 &	37.07 
 \\
\specialrule{0.07em}{0pt}{0pt}

\multirow{6}{*}{\makecell{\emph{\textbf{Dynamic}}\\ \emph{\textbf{Process}}}} 
& AC &   & \underline{91.92} &	\textbf{92.69} &	70.77 &	44.62 & &	\textbf{71.92} &	5.00 &	34.62 &	51.54 &		\underline{58.08} &	12.31 &	46.92 &	20.38 &	41.92 &	30.38 
 \\
& PQ  &  & \underline{94.62} &	\textbf{97.31} &	93.46 &	70.00 & &		\underline{85.77} &	55.00 &	\textbf{88.46} &	64.62 &	76.15 &	36.92 &	79.23 &	48.05 &	65.38 &	71.54 
 \\
& SC  &  & \textbf{71.92} &	\underline{62.69} &	36.15 &	23.08 & &	28.85 &	3.85 &	8.08 &	\textbf{31.15} &		\underline{29.62} &	7.69 &	23.08 &	9.23 &	20.38 &	3.85 
 \\
& LC  &  & \textbf{70.38} &	\underline{64.62} &	36.92 &	21.54 & &		\underline{30.38} &	4.23 &	18.08 &	30.00 &	\textbf{35.38} &	9.23 &	28.46 &	10.77 &	28.85 &	11.54 
 \\
& SP &   & \underline{70.38} &	\textbf{71.92} &	43.85 &	38.46 & &	\textbf{47.69} &	4.62 &	21.15 &	34.23 &		\underline{35.38} &	14.62 &	31.54 &	16.54 &	31.15 &	14.62 
 \\
& Avg &  & \textbf{79.85} &	\underline{77.85} &	56.23 &	39.54 & &	\textbf{52.92} &	14.54 &	34.08 &	42.31 &		\underline{46.92} &	16.15 &	41.85 &	20.99 &	37.54 &	26.38 
 \\
\specialrule{0.07em}{0pt}{0pt}

\multirow{6}{*}{\makecell{\emph{\textbf{Scientific}}\\ \emph{\textbf{Simulation}}}} 
& AC  &  & \textbf{94.57} &		\underline{86.96} &	66.30 &	33.70 & &	\textbf{55.43} &	5.43 &	30.43 &	43.48 &	\underline{48.91} &	7.61 &	36.96 &	8.70 &	39.13 &	29.35 
 \\
& PQ  &  & \textbf{93.48} &		\underline{90.22} &	80.43 &	70.65 & &	73.91 &	47.83 &	\underline{75.00} &	66.30 &	\textbf{80.43} &	33.70 &	73.91 &	42.39 &	50.00 &	\underline{75.00} 
 \\
& SC  &  & \textbf{72.83} &		\underline{68.48} &	33.70 &	20.65 & &	\textbf{26.09} &	0.00 &	8.70 &	\underline{25.00} &	\textbf{26.09} &	4.35 &	20.65 &	3.26 &	21.74 &	3.26 
 \\
& LC &   & \textbf{78.26} &		\underline{73.91} &	32.61 &	17.39 & &	\underline{28.26} &	2.17 &	16.30 &	27.17 &	\textbf{31.52} &	4.35 &	17.39 &	4.35 &	25.00 &	4.35 
 \\
& SP &   & 	\underline{73.91} &	\textbf{78.26} &	43.48 &	25.00 & &	\textbf{36.96} &	6.52 &	26.09 &	\underline{33.70} &	30.43 &	11.96 &	22.83 &	10.87 &	\underline{33.70} &	9.78 
 \\
& Avg &  & \textbf{82.61} &		\underline{79.57} &	51.30 &	33.48 & &	\textbf{44.13} &	12.39 &	31.30 &	39.13 &	\underline{43.48} &	12.39 &	34.35 &	13.91 &	33.91 &	24.35 
 \\


\specialrule{0.1em}{0pt}{0pt}

\end{tabular}
}
\end{table*}

\begin{figure*}[th!]
\centering
\includegraphics[width=0.8\textwidth]{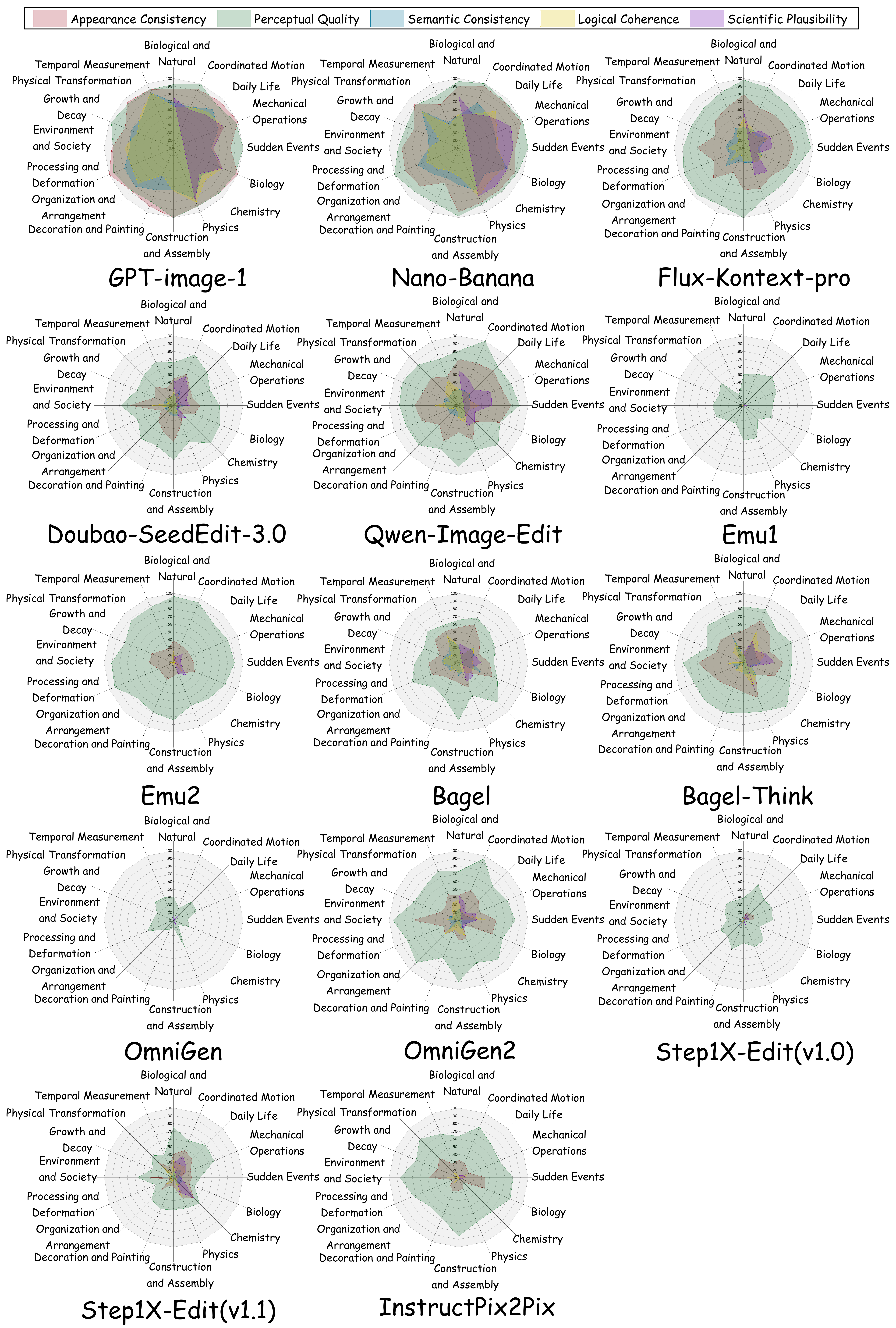} 
\caption{The average scores for 14 models across 16 subtasks.}
\label{APPENDIX-RADAR-ALL}
\end{figure*}

Tab.~\ref{Specific Scores Across Four Four Fundamental Tasks} and Fig.~\ref{APPENDIX-RADAR-ALL} show the specific scores of 14 models across 4 fundamental tasks and 16 sub-tasks, respectively. Compared to open-source models, proprietary models exhibit significantly more balanced performance. In all tasks, the models particularly excel in the perceptual quality assessment dimension, demonstrating their ability to generate natural, smooth, and high-quality images, avoiding common issues such as distortion and blurring. This result indicates that current models effectively address challenges related to image quality when generating visual content.

Most models score slightly lower in the appearance consistency dimension compared to the perceptual quality dimension, yet still demonstrate considerable capability. However, some models have still failed to effectively adapt to the new paradigm of intermediate logic path editing, resulting in poor performance in the appearance consistency dimension. For example, models like Emu1 and OmniGen encounter significant obstacles in this dimension, with performance far below that of other models.

There are significant variations in performance across models in terms of semantic consistency and logical consistency. Except for GPT-Image-1 and Nano-Banana, other models show significant imbalances in these two dimensions, with a noticeable drop in scores. Notably, models like Emu1 and OmniGen score almost zero in both semantic consistency and logical consistency, highlighting the limitations of current models in handling complex logical relationships.

Among the open-source models, all except Qwen-Image-Edit, Bagel, and Bagel-Think show relatively poor performance. Among them, a few models, such as OmniGen2 and Step1X-Edit(v1.1), show slight improvements in certain sub-tasks, achieving some scores. However, these advancements are not applicable to a broader range of tasks. Overall, the performance of open-source models still lags significantly behind that of proprietary models, with notable gaps in multiple key dimensions.

\subsection{Accuracy of Models Across 16 Sub-Tasks}

\setlength{\tabcolsep}{5pt}
\begin{table*}[t]
\renewcommand\arraystretch{1.2}
\caption{Accuracy performance of different models across 16 sub-tasks, including State Transition: Construction and Assembly (CA), Decoration and Painting (DP), Organization and Arrangement (OA), Processing and Deformation (PD). Temporal Sequence: Environment and Society (ES), Growth and Decay (GD), Physical Transformation (PT), Temporal Measurement (TM). Dynamic Process: Biology and Nature (BN), Coordinated Motion (CM), Daily Life (DL), Mechanical Operations (MO), Sudden Events (SE). Scientific Simulation: Biology (BI), Chemistry (CH), Physics (PH). The performance of open-source and proprietary models is separately marked, with the best performance in \textbf{bold} and the second-best performance \underline{underlined}.}
\label{Accuracy of Models Across 16 Sub-Tasks}
\centering
\resizebox{\textwidth}{!}{
\begin{tabular}{cccccccccccccccccc}
\specialrule{0.1em}{0pt}{2pt}
\multicolumn{2}{c}{\multirow{2}{*}[-7ex]{SubTasks}} & & \multicolumn{4}{c}{\textbf{Proprietary Models}} & & \multicolumn{10}{c}{\textbf{Open-Source Models}} \\ 
\cmidrule{4-7} \cmidrule{9-18}
& & & \multicolumn{1}{c}{\rotatebox{90}{\small GPT-Image-1}} &
\multicolumn{1}{c}{\rotatebox{90}{\small Nano-Banana}} &
\multicolumn{1}{c}{\rotatebox{90}{\small Flux-Kontext-pro}} &
\multicolumn{1}{c}{\rotatebox{90}{\small Doubao-SeedEdit-3.0}} & &
\multicolumn{1}{c}{\rotatebox{90}{\small Qwen-Image-Edit}} &
\multicolumn{1}{c}{\rotatebox{90}{\small Emu1}} &
\multicolumn{1}{c}{\rotatebox{90}{\small Emu2}} &
\multicolumn{1}{c}{\rotatebox{90}{\small Bagel}} &
\multicolumn{1}{c}{\rotatebox{90}{\small Bagel-Think}} &
\multicolumn{1}{c}{\rotatebox{90}{\small OmniGen}} &
\multicolumn{1}{c}{\rotatebox{90}{\small OmniGen2}} &
\multicolumn{1}{c}{\rotatebox{90}{\small Step1X-Edit (v1.0)}} &
\multicolumn{1}{c}{\rotatebox{90}{\small Step1X-Edit (v1.1)}} &
\multicolumn{1}{c}{\rotatebox{90}{\small InstructPix2Pix}} \\
\specialrule{0.07em}{0pt}{0pt}
\multirow{5}{*}{\makecell{\emph{\textbf{State}}\\ \emph{\textbf{Transition}}}}
& CA  &  & \textbf{14.29} &	\underline{7.14} &	0.00 &	0.00 & &	0.00 &	0.00 &	0.00 &	0.00 &	0.00 &	0.00 &	0.00 &	0.00 &	0.00 &	0.00 \\

& DP &   & \textbf{8.33} &	\textbf{8.33} &	0.00 &	0.00 & &	0.00 &	0.00 &	0.00 &	0.00 &	0.00 &	0.00 &	0.00 &	0.00 &	0.00 &	0.00  \\

& OA  &  & \textbf{40.00} &	\underline{10.00} &	\underline{10.00} &	0.00 & &	0.00 &	0.00 &	0.00 &	0.00 &	0.00 &	0.00 &	0.00 &	0.00 &	0.00 &	0.00 \\

& PD  &  & \underline{7.69} &	\textbf{15.38} &	0.00 &	0.00 & &	0.00 &	0.00 &	0.00 &	0.00 &	0.00 &	0.00 &	0.00 &	0.00 &	0.00 &	0.00 \\

& Avg & & \textbf{16.33} &	\underline{10.20} &	2.04 &	0.00 & &	0.00 &	0.00 &	0.00 &	0.00 &	0.00 &	0.00 &	0.00 &	0.00 &	0.00 &	0.00 \\
\specialrule{0.07em}{0pt}{0pt}

\multirow{5}{*}{\makecell{\emph{\textbf{Temporal}}\\ \emph{\textbf{Sequence}}}} 
& ES  &  & \underline{10.53} &	\textbf{15.79} &	0.00 &	0.00 & &	0.00 &	0.00 &	0.00 &	0.00 &	0.00 &	0.00 &	\textbf{5.26} &	0.00 &	0.00 &	0.00 \\

& GD  &  & \textbf{20.00} &	0.00 &	0.00 &	0.00 & &	0.00 & 	0.00 &	0.00 &	0.00 &	0.00 &	0.00 &	0.00 &	0.00 &	0.00 &	0.00 \\

& PT  &  & \underline{12.00} &	\textbf{40.00} &	0.00 &	0.00 & &	0.00 &	0.00 &	0.00 &	0.00 &	0.00 &	0.00 &	0.00 &	0.00 &	0.00 &	0.00 \\

& TM  &  & \textbf{57.14} &	0.00 &	0.00 &	0.00 & &	0.00 &	0.00 &	0.00 &	0.00 &	\textbf{14.29} &	0.00 &	0.00 &	0.00 &	0.00 &	0.00 \\

& Avg & & \underline{18.18} &	\textbf{19.70} &	0.00 &	0.00 & &	0.00 &	0.00 &	0.00 &	0.00 &	\textbf{1.52} &	0.00 &	\textbf{1.52} &	0.00 &	0.00 &	0.00 \\
\specialrule{0.07em}{0pt}{0pt}
\multirow{6}{*}{\makecell{\emph{\textbf{Dynamic}}\\ \emph{\textbf{Process}}}} 
& BN &   & \textbf{15.38} &	\underline{7.69} &	0.00 &	0.00 & &	0.00 &	0.00 &	0.00 &	0.00 &	0.00 &	0.00 &	0.00 &	0.00 &	0.00 &	0.00 \\

& CM  &  & 0.00 &	\textbf{7.69} &	0.00 &	0.00 & &	\textbf{7.69} &	0.00 &	0.00 &	0.00 &	\textbf{7.69} &	0.00 &	0.00 &	0.00 &	0.00 &	0.00 \\

& DL  &  & \textbf{28.57} &	\underline{9.52} &	4.76 &	0.00 & &	0.00 &	0.00 &	0.00 &	0.00 &	0.00 &	0.00 &	0.00 &	0.00 &	0.00 &	0.00 \\

& MO  &  & \textbf{22.22} &	\textbf{22.22} &	0.00 &	0.00 & &	0.00 &	0.00 &	0.00 &	0.00 &	0.00 &	0.00 &	0.00 &	0.00 &	0.00 &	0.00 \\

& SE &  & \textbf{11.11} &	0.00 &	0.00 &	0.00 & &	0.00 &	0.00 &	0.00 &	0.00 &	0.00 &	0.00 &	0.00 &	0.00 &	0.00 &	0.00 \\

& Avg & & \textbf{16.92} &	\underline{9.23} &	1.54 &	0.00 & &	\textbf{1.54} &	0.00 &	0.00 &	0.00 &	\textbf{1.54} &	0.00 &	0.00 &	0.00 &	0.00 &	0.00 \\
\specialrule{0.07em}{0pt}{0pt}
\multirow{4}{*}{\makecell{\emph{\textbf{Scientific}}\\ \emph{\textbf{Simulation}}}} 
& BI  &  & 0.00 &	\textbf{28.57} &	0.00 &	0.00 & &	0.00 &	0.00 &	0.00 &	0.00 &	0.00 &	0.00 &	0.00 &	0.00 &	0.00 &	0.00 
 \\

& CH  &  & 0.00 &	0.00 &	0.00 &	0.00 & &	0.00 &	0.00 &	0.00 &	0.00 &	0.00 &	0.00 &	0.00 &	0.00 &	0.00 &	0.00 
 \\

& PH  &  & \textbf{33.33} &	\underline{11.11} &	0.00 &	0.00 & &	0.00 &	0.00 &	0.00 &	0.00 &	0.00 &	0.00 &	0.00 &	0.00 &	0.00 &	0.00 \\

& Avg &  & \textbf{13.04} &	\textbf{13.04} &	0.00 &	0.00 & &	0.00 &	0.00 &	0.00 &	0.00 &	0.00 &	0.00 &	0.00 &	0.00 &	0.00 &	0.00 \\
\specialrule{0.07em}{0pt}{0pt}
\multicolumn{2}{c}{\textbf{Overall Accuracy}} & & \textbf{16.75} & \underline{13.30} & 0.99 & 0.00 & & \underline{0.49} & 0.00 & 0.00 & 0.00 & \textbf{0.99} & 0.00 & \underline{0.49} & 0.00 & 0.00 & 0.00\\

\specialrule{0.1em}{0pt}{0pt}

\end{tabular}
}
\label{table3}
\end{table*}

Tab.~\ref{Accuracy of Models Across 16 Sub-Tasks} shows the accuracy scores of each model across 16 sub-tasks. The effective scores are primarily concentrated in GPT-Image-1 and Nano-Banana. Both models perform well in multiple sub-tasks. Although GPT-Image-1 outperforms Nano-Banana overall, Nano-Banana still has an advantage in certain sub-tasks. In the case of open-source models, all models have an accuracy of 0\% in the state transition and scientific simulation category tasks, and in tasks from other categories, only a few models show slight improvements in their scores. Overall, even the most advanced models achieve only 16.75\% accuracy, with more than half of the models scoring 0\%. This indicates that current models are still in the early stages of solving intermediate logic path editing tasks, far from meeting the requirements for practical application.

\begin{figure*}[th!]
\centering
\includegraphics[width=0.8\textwidth]{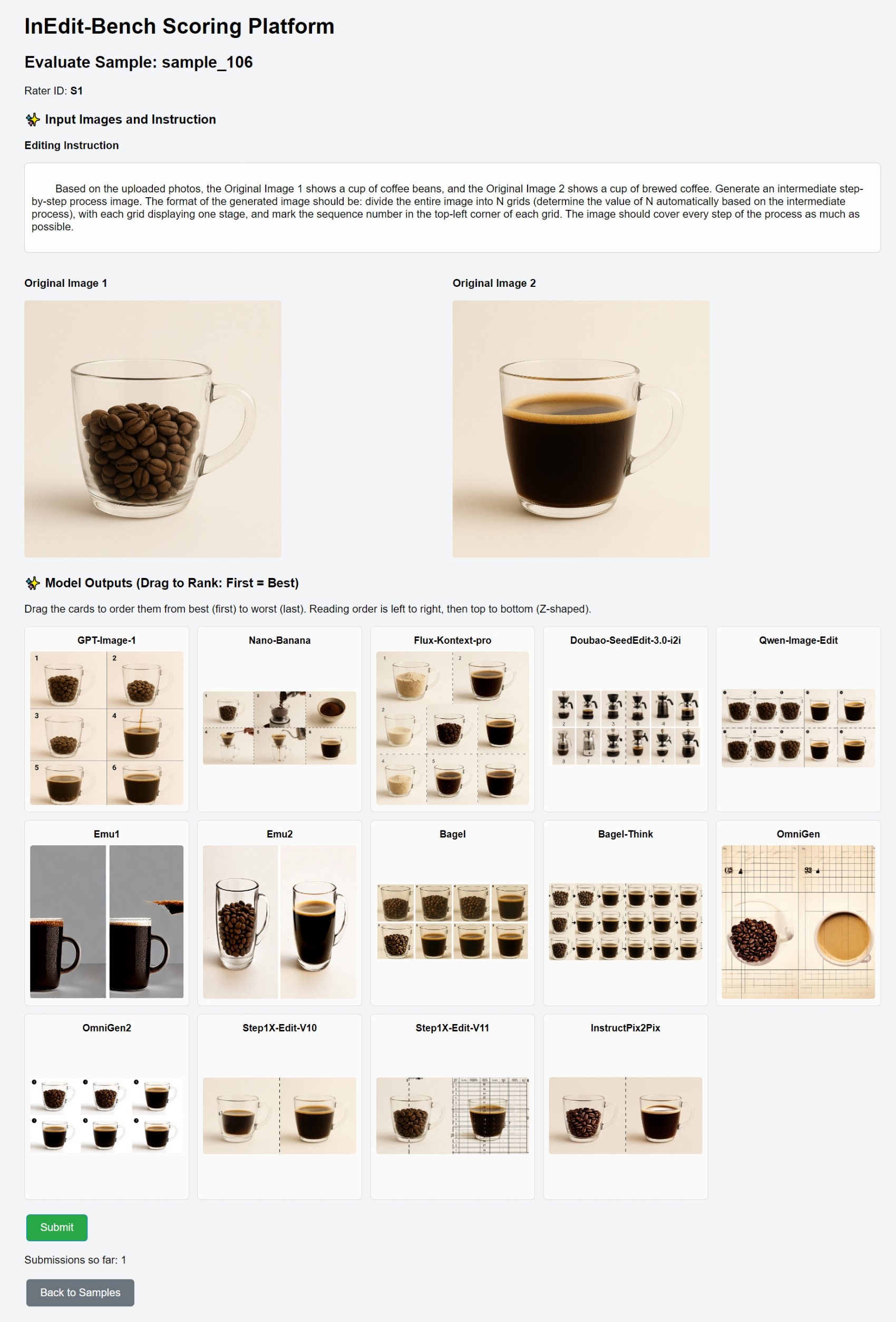} 
\caption{The instance evaluation interface provided for human evaluators.}
\label{human_evaluator}
\end{figure*}


\section{Human Evaluator} \label{Human Evaluator}
We enhanced the robustness of the evaluation system by integrating a human evaluation mechanism. Specifically, we designed and implemented an interactive manual evaluation platform that enables users to intuitively perceive and compare the generation results of different models, and to perform preference ranking of the model outputs based on their subjective judgment. The key modules of the evaluation platform include: Evaluator Identity Setting, Evaluation Instance Selection, and the Instance Evaluation Interface. As illustrated in Fig.~\ref{human_evaluator}, the instance evaluation interface displays the original image, the editing instruction, and the collection of output results from different models. Evaluators are able to rank the models using a drag-and-drop function. The resulting evaluations will be submitted to the backend server in the form of structured data for later quantitative analysis.


\section{Data Source of InEdit-Bench} \label{Data Source of InEdit-Bench}
Input images for the InEdit-Bench dataset are primarily sourced from the following categories:

(1) Images generated by image generation models.

(2) Images derived from existing datasets and benchmarks.

(3) Images collected from the internet under permissive licenses.

\section{Limitations} \label{Limitations}

This study aims to establish a pioneering benchmark for intermediate logical reasoning and multi-step editing tasks. However, as an initial exploration, the current benchmark still has several aspects that require improvement. We openly acknowledge its potential limitations, such as the dataset’s insufficient scale to cover all complex scenarios and the task categorization that may not exhaust all possibilities. Future work will focus on addressing these issues to build a more comprehensive and robust benchmark.


\section{Representative Example Images from InEdit-Bench}
\label{Representative Example Images from InEdit-Bench}
\begin{figure*}[t!]
\centering
\includegraphics[width=0.9\textwidth]{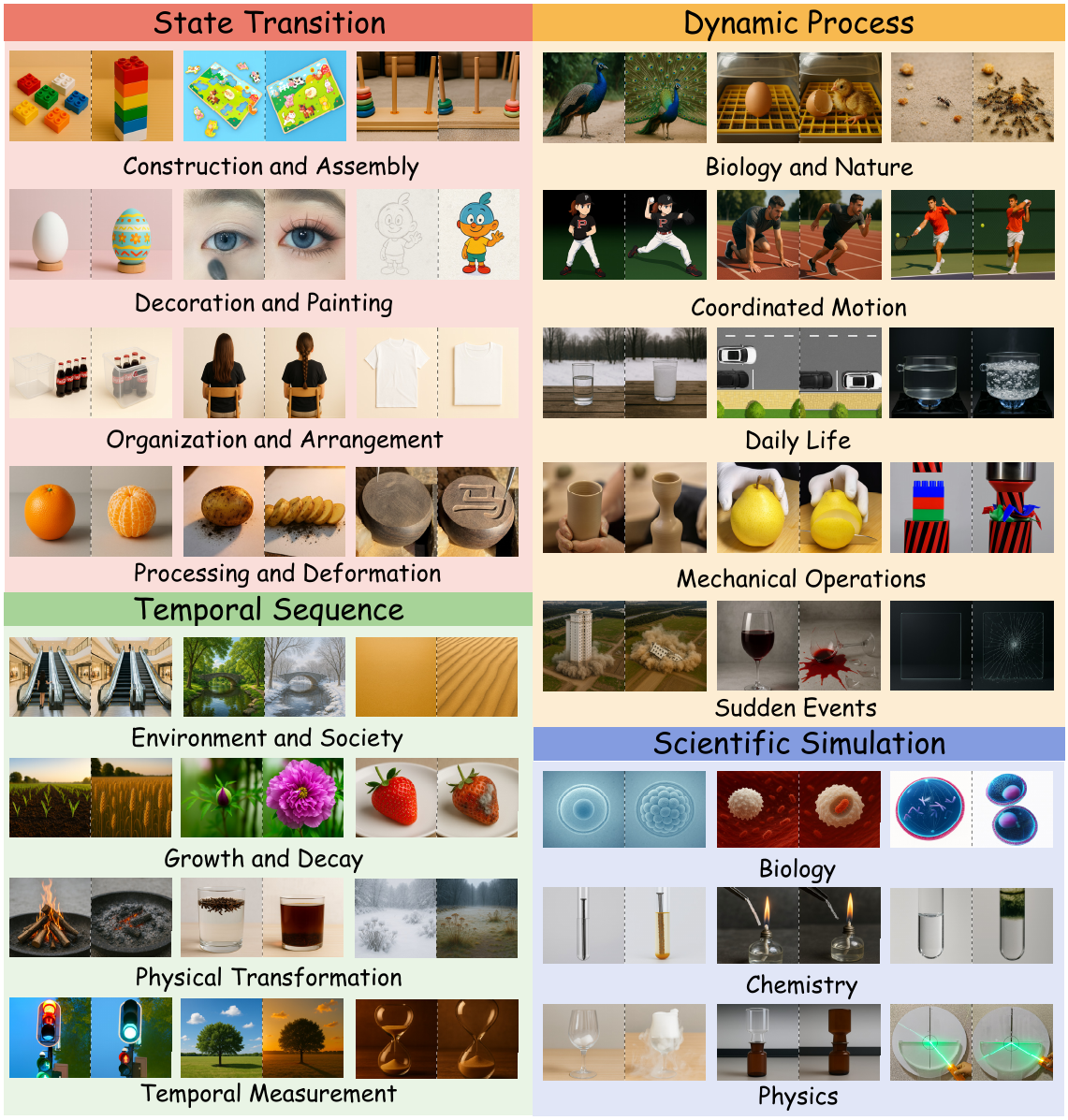} 
\caption{Representative Examples of InEdit-Bench Subtasks}
\label{subtask_data10142209}

\end{figure*}

In this section, we present representative example images from the 16 subtasks in InEdit-Bench, with each subtask corresponding to a distinct testing scenario. Fig.~\ref{subtask_data10142209} illustrate examples from the four task categories: 4 subtasks of state transition, 4 subtasks of temporal sequence, 5 subtasks of dynamic process, and 3 subtasks of scientific simulation.

\section{Detailed Outputs of Evaluated Models} \label{Detailed Outputs of Evaluated Models}
\begin{figure*}[t!]
\centering
\includegraphics[width=0.75\textwidth]{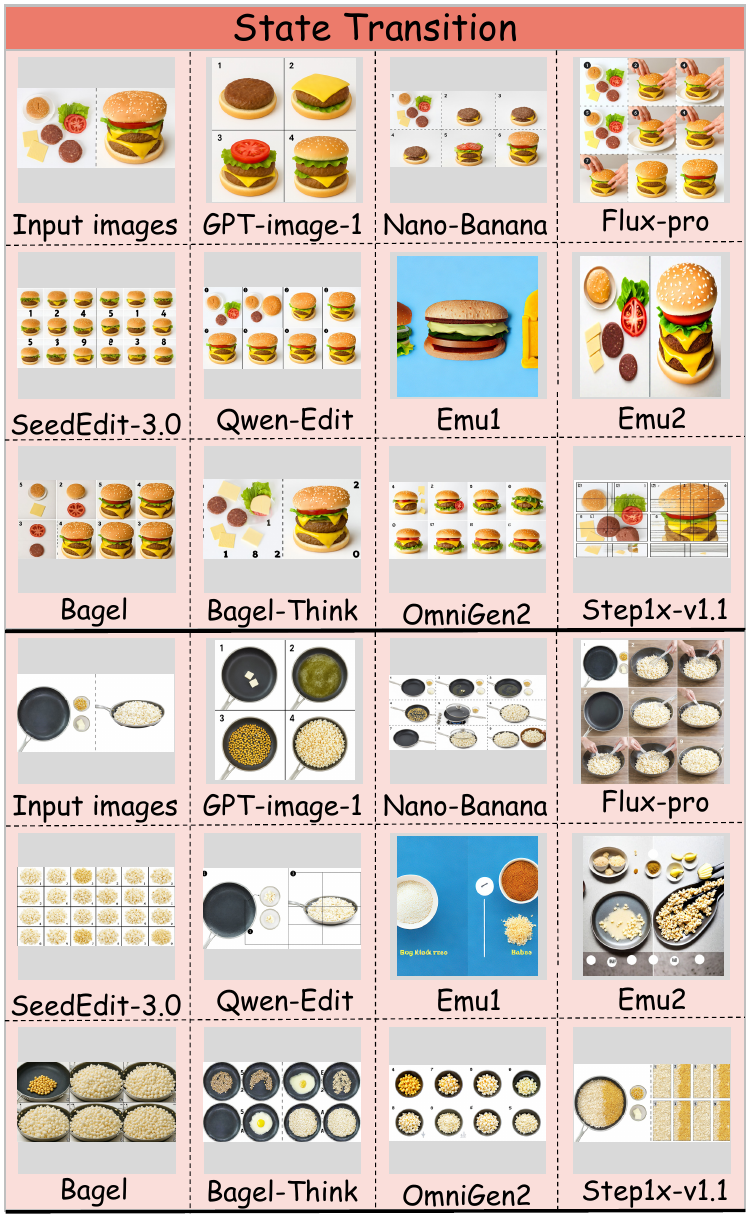} 
\caption{State Transition Outputs - Part1.}
\label{figstate_part1}
\end{figure*}

\begin{figure*}[t!]
\centering
\includegraphics[width=0.75\textwidth]{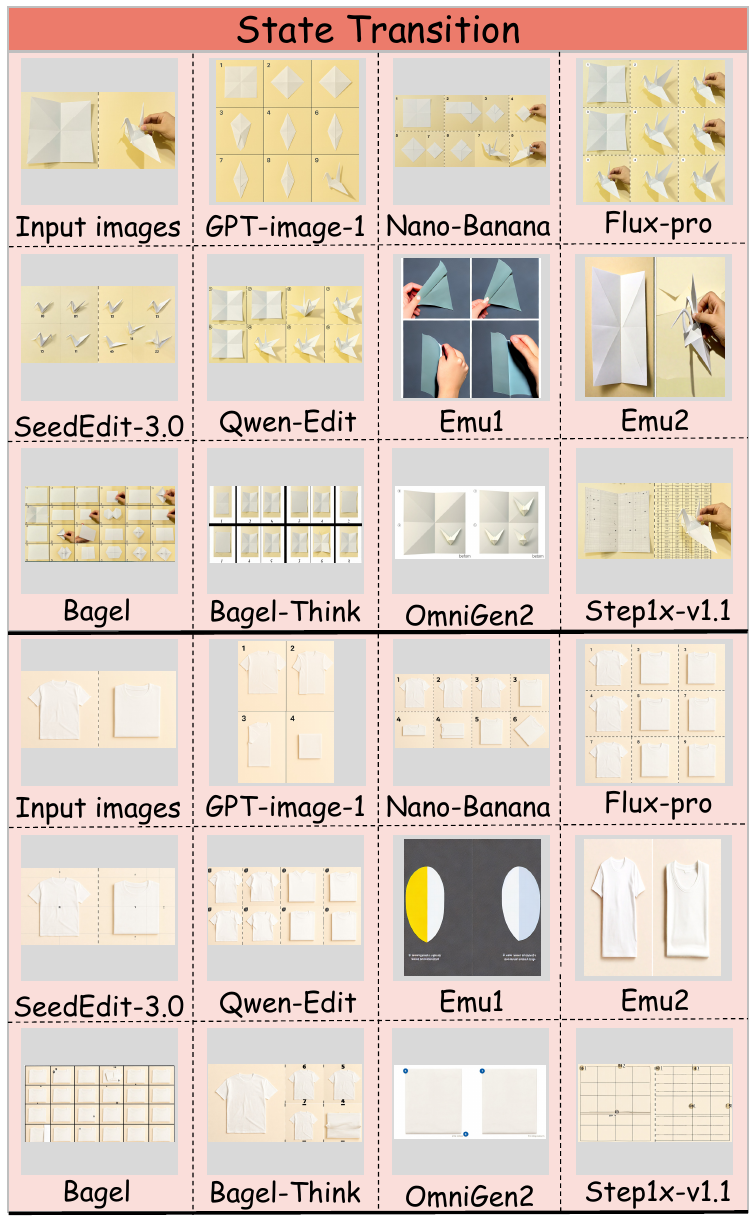} 
\caption{State Transition Outputs - Part2.}
\label{figstate_part2}
\end{figure*}

\begin{figure*}[t!]
\centering
\includegraphics[width=0.75\textwidth]{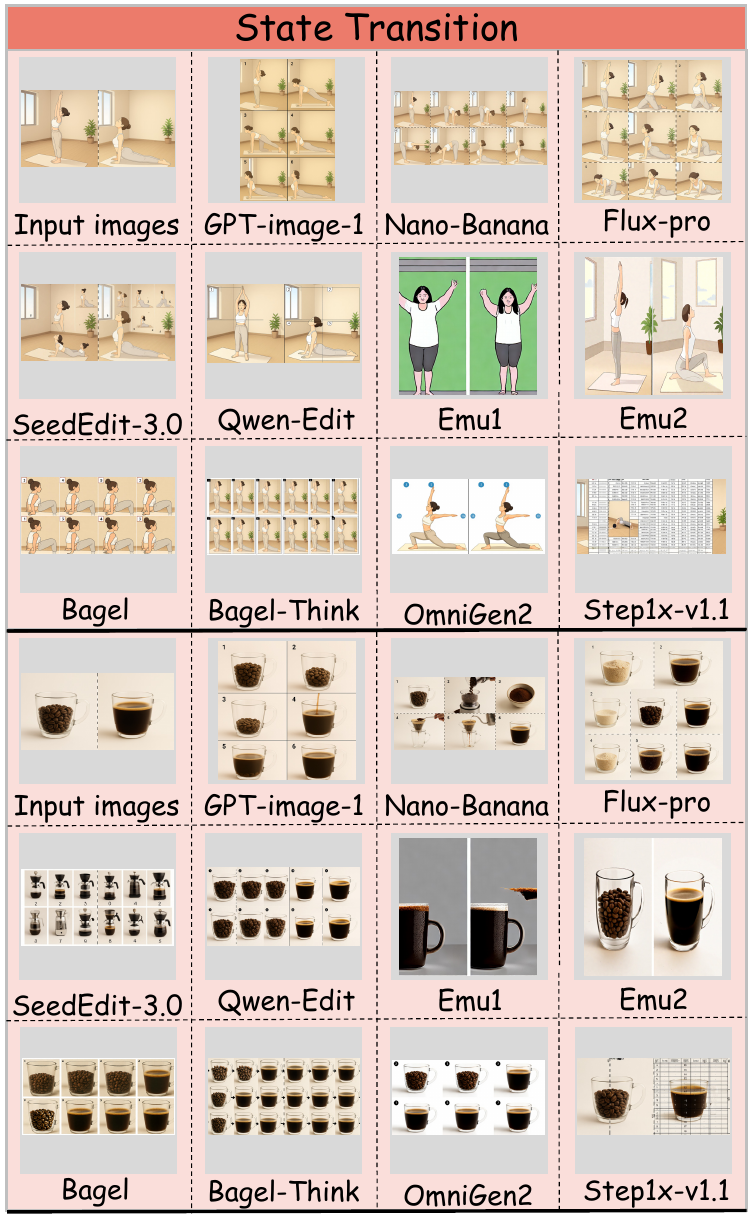} 
\caption{State Transition Outputs - Part3.}
\label{figstate_part3}
\end{figure*}

\begin{figure*}[t!]
\centering
\includegraphics[width=0.75\textwidth]{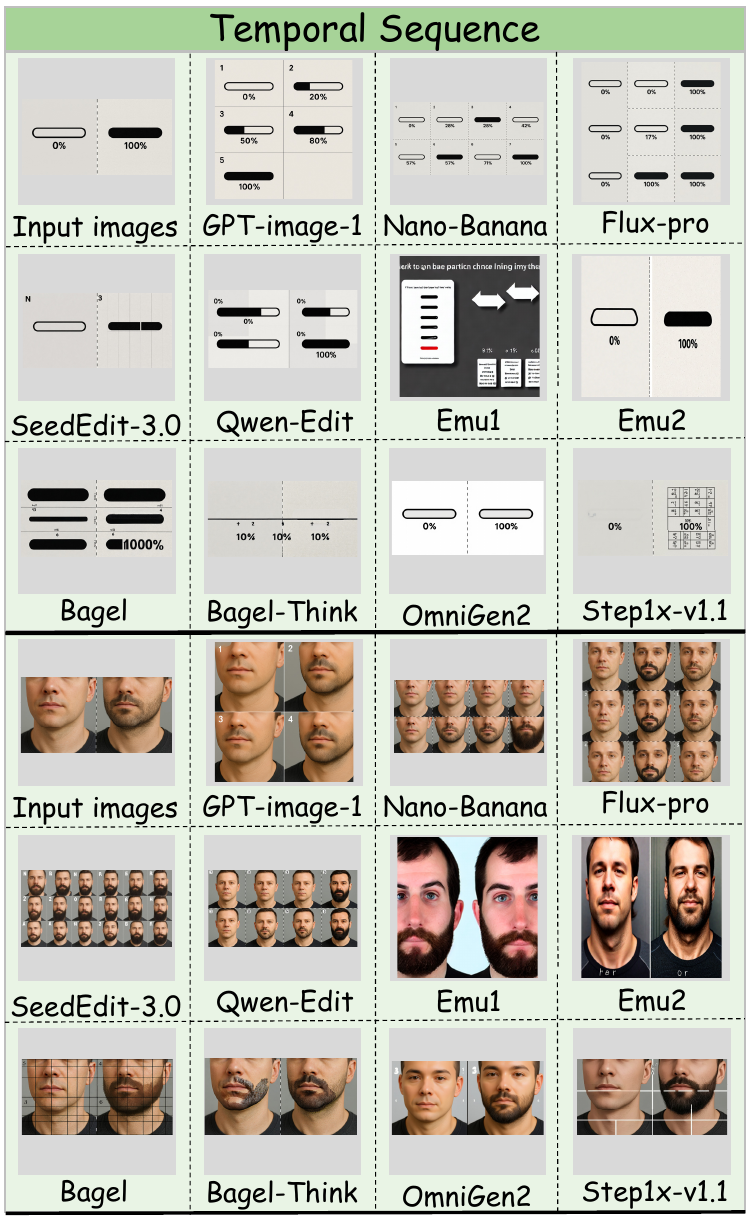} 
\caption{Temporal Sequence Outputs - Part1.}
\label{figTemporal_part1}
\end{figure*}

\begin{figure*}[t!]
\centering
\includegraphics[width=0.75\textwidth]{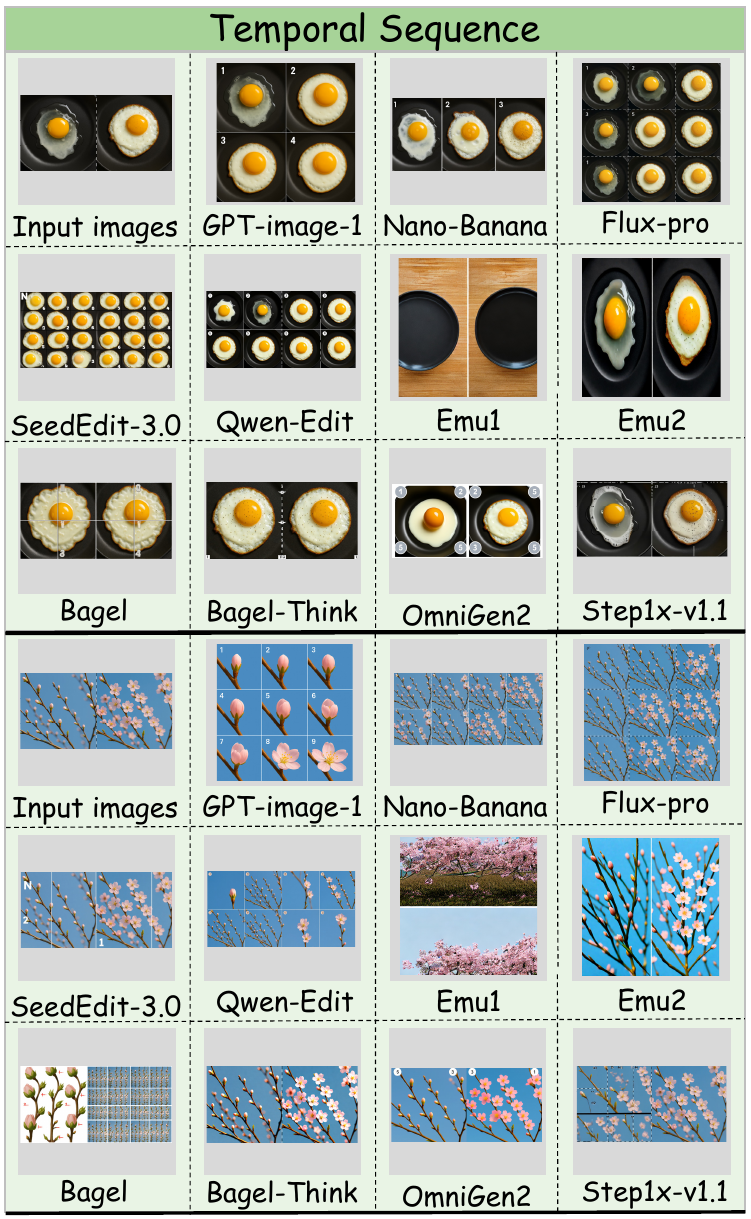} 
\caption{Temporal Sequence Outputs - Part2.}
\label{figTemporal_part2}
\end{figure*}

\begin{figure*}[t!]
\centering
\includegraphics[width=0.75\textwidth]{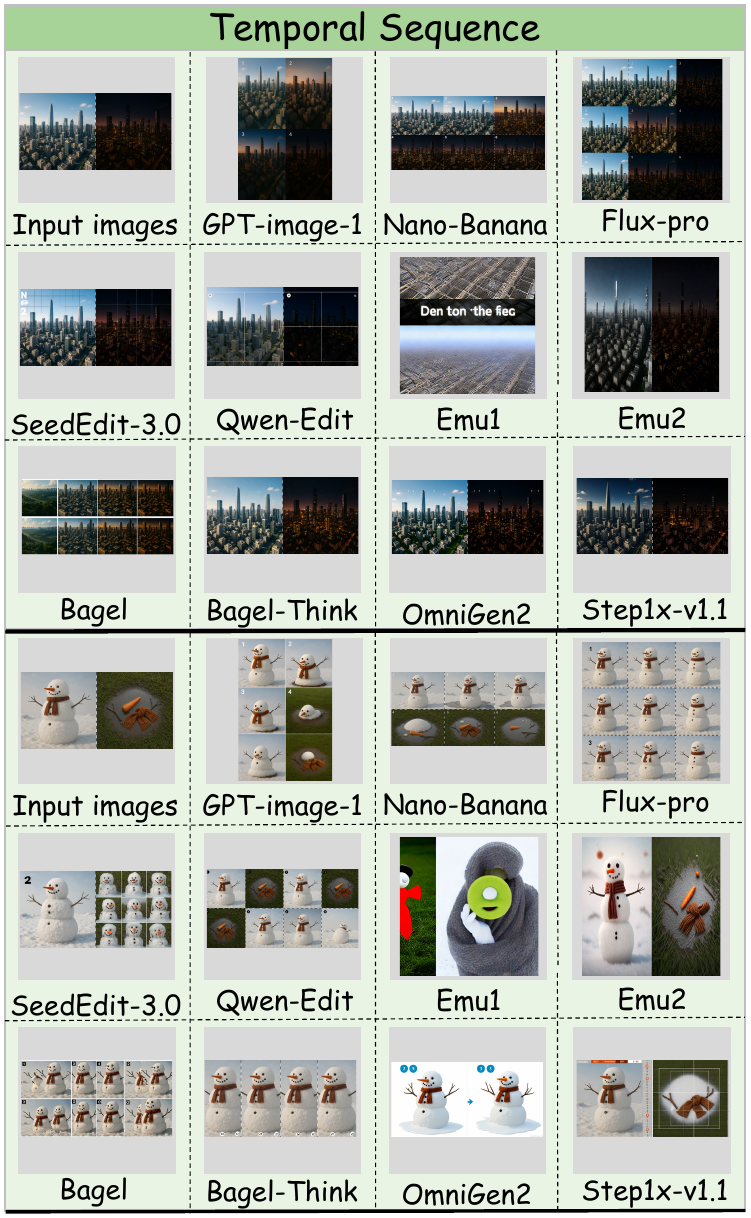} 
\caption{Temporal Sequence Outputs - Part3.}
\label{figTemporal_part3}
\end{figure*}

\begin{figure*}[t!]
\centering
\includegraphics[width=0.75\textwidth]{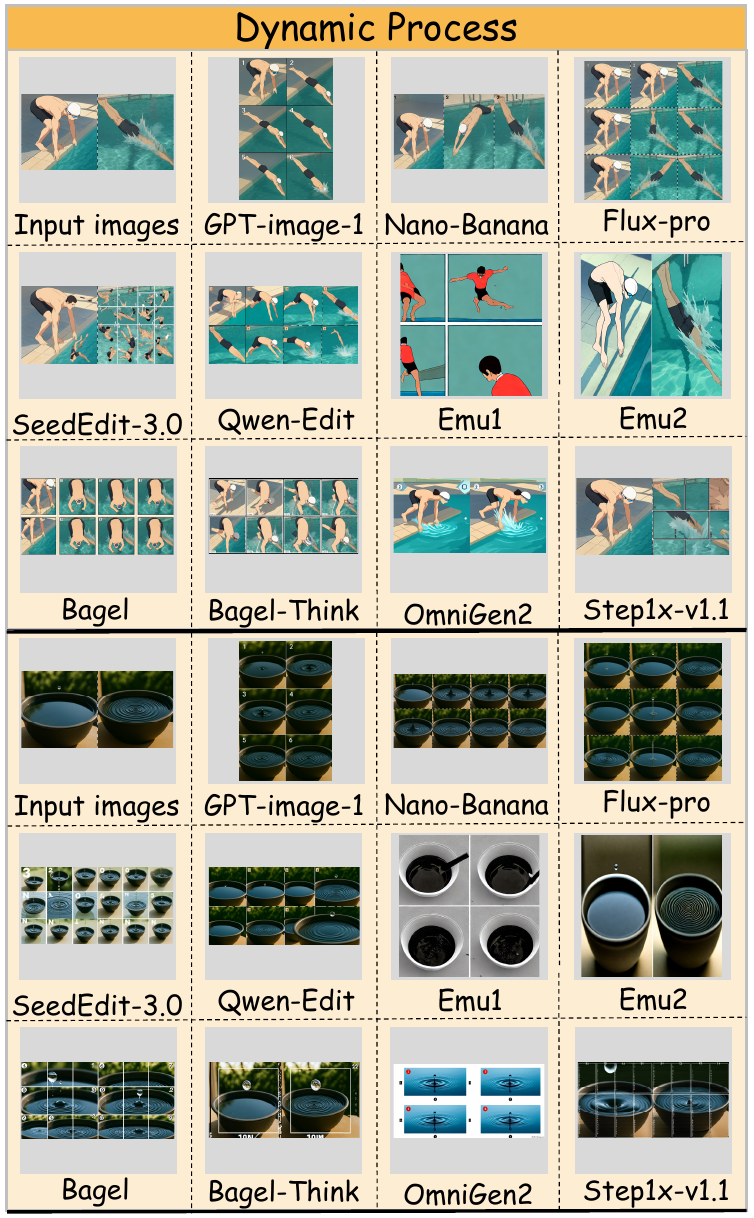} 
\caption{Dynamic Process Outputs - Part1.}
\label{figdynamic_part1}
\end{figure*}

\begin{figure*}[t!]
\centering
\includegraphics[width=0.75\textwidth]{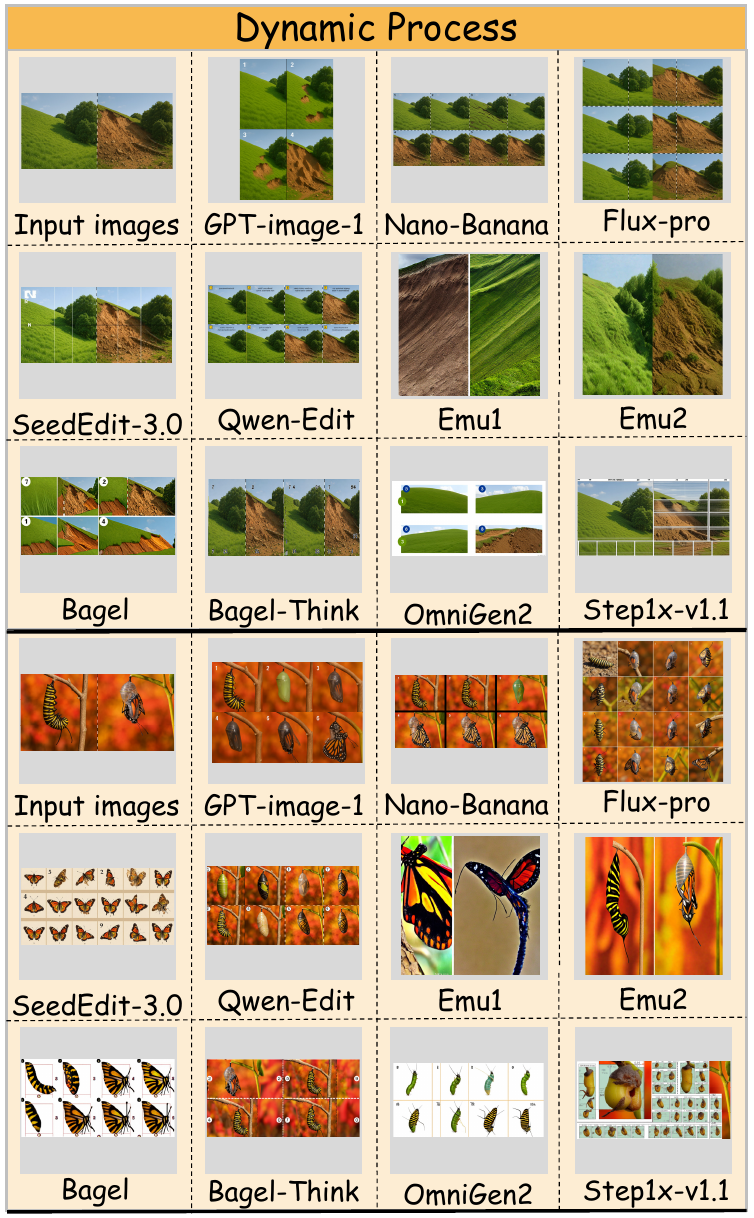} 
\caption{Dynamic Process Outputs - Part2.}
\label{figdynamic_part2}
\end{figure*}

\begin{figure*}[t!]
\centering
\includegraphics[width=0.75\textwidth]{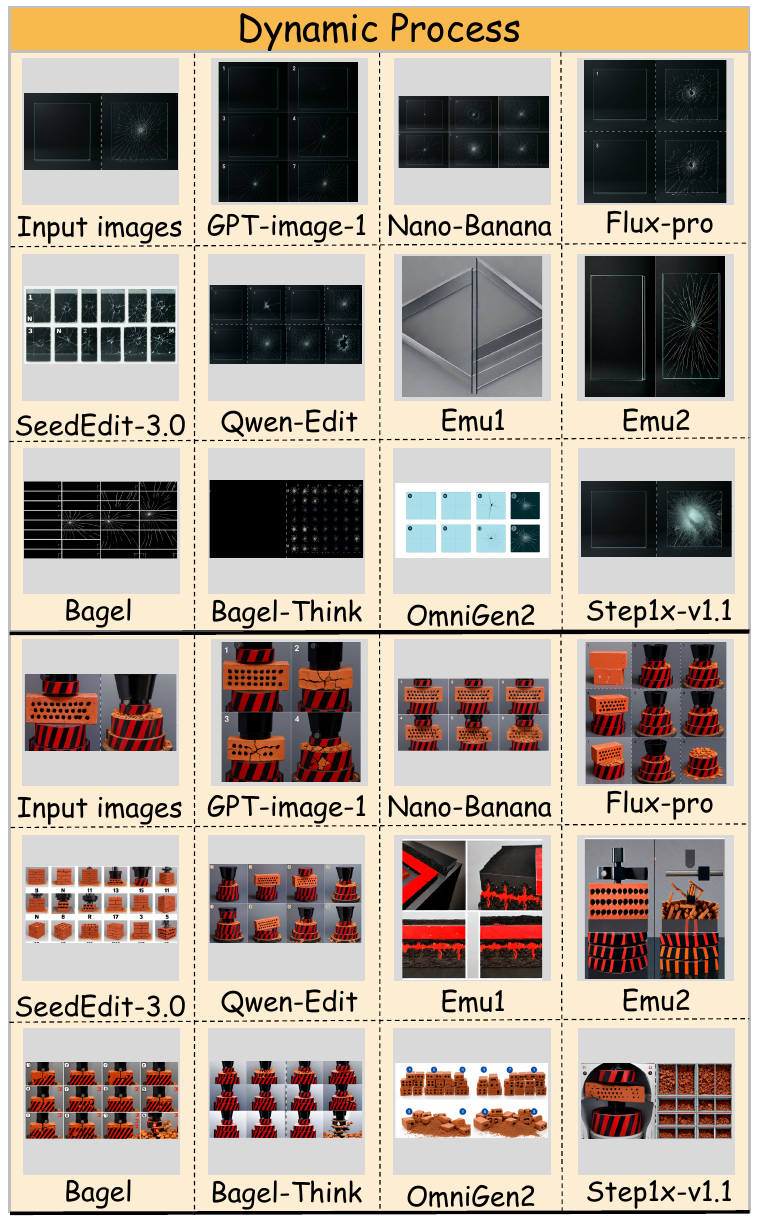} 
\caption{Dynamic Process Outputs - Part3.}
\label{figdynamic_part3}
\end{figure*}

\begin{figure*}[t!]
\centering
\includegraphics[width=0.75\textwidth]{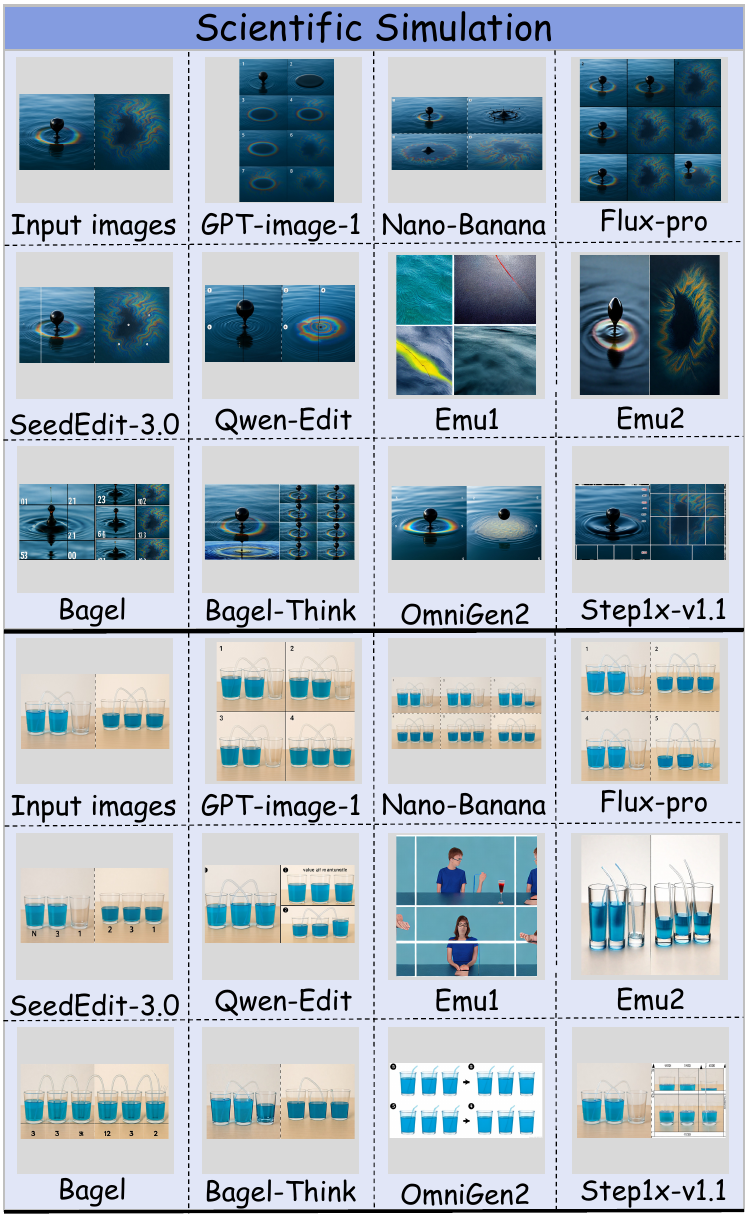} 
\caption{Scientific Simulation Outputs - Part1.}
\label{figsimulation_part1}
\end{figure*}

\begin{figure*}[t!]
\centering
\includegraphics[width=0.75\textwidth]{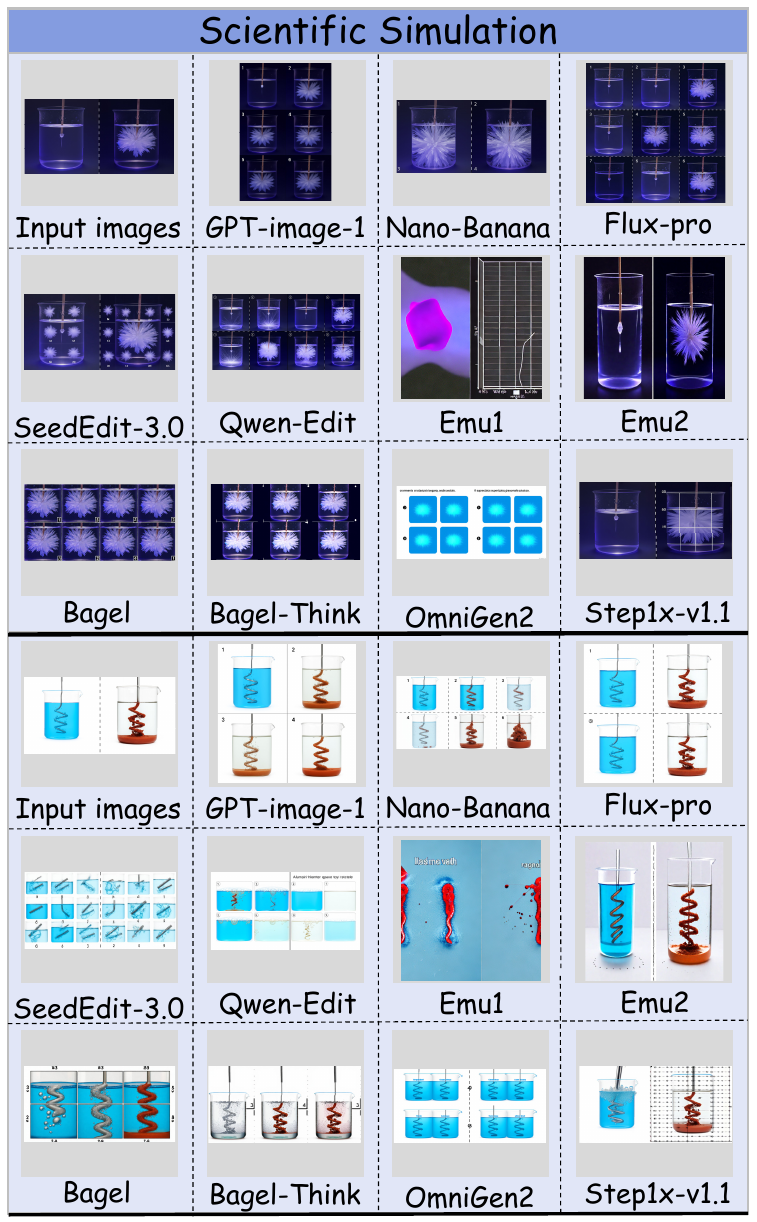} 
\caption{Scientific Simulation Outputs - Part2.}
\label{figsimulation_part2}
\end{figure*}

\begin{figure*}[t!]
\centering
\includegraphics[width=0.75\textwidth]{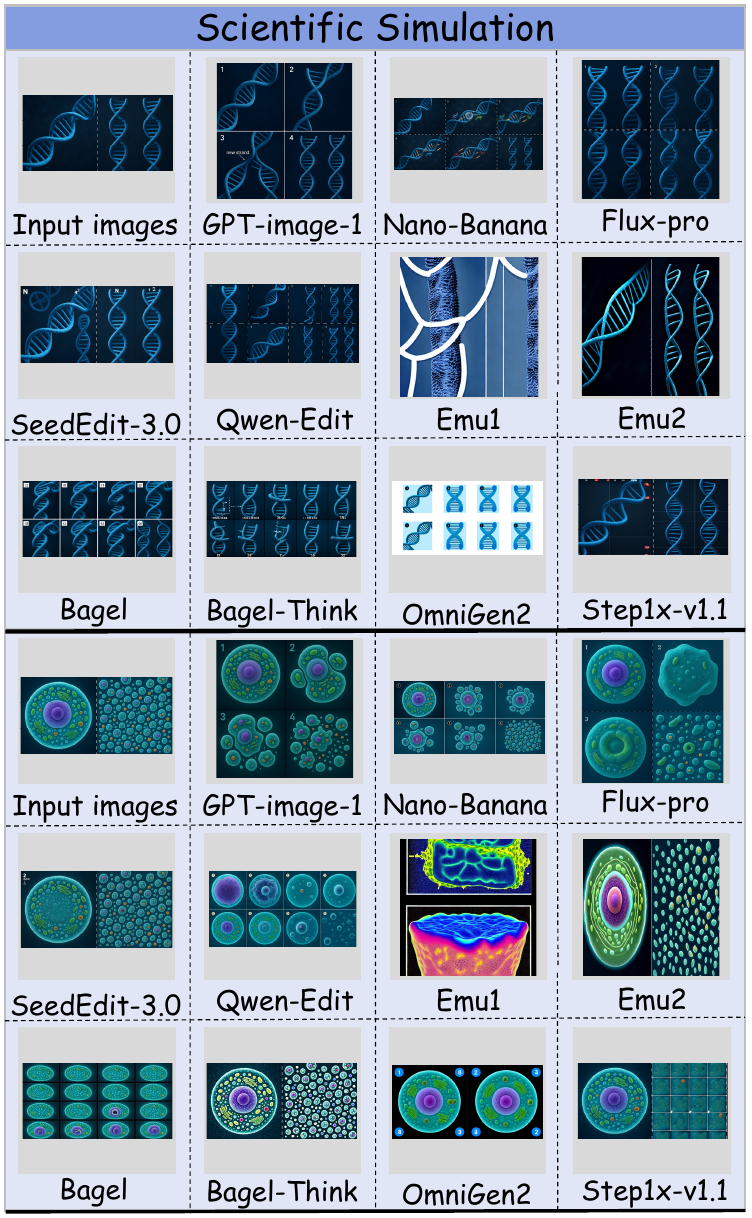} 
\caption{Scientific Simulation Outputs - Part3.}
\label{figsimulation_part3}
\end{figure*}

\begin{figure*}[t!]
\centering
\includegraphics[width=0.75\textwidth]{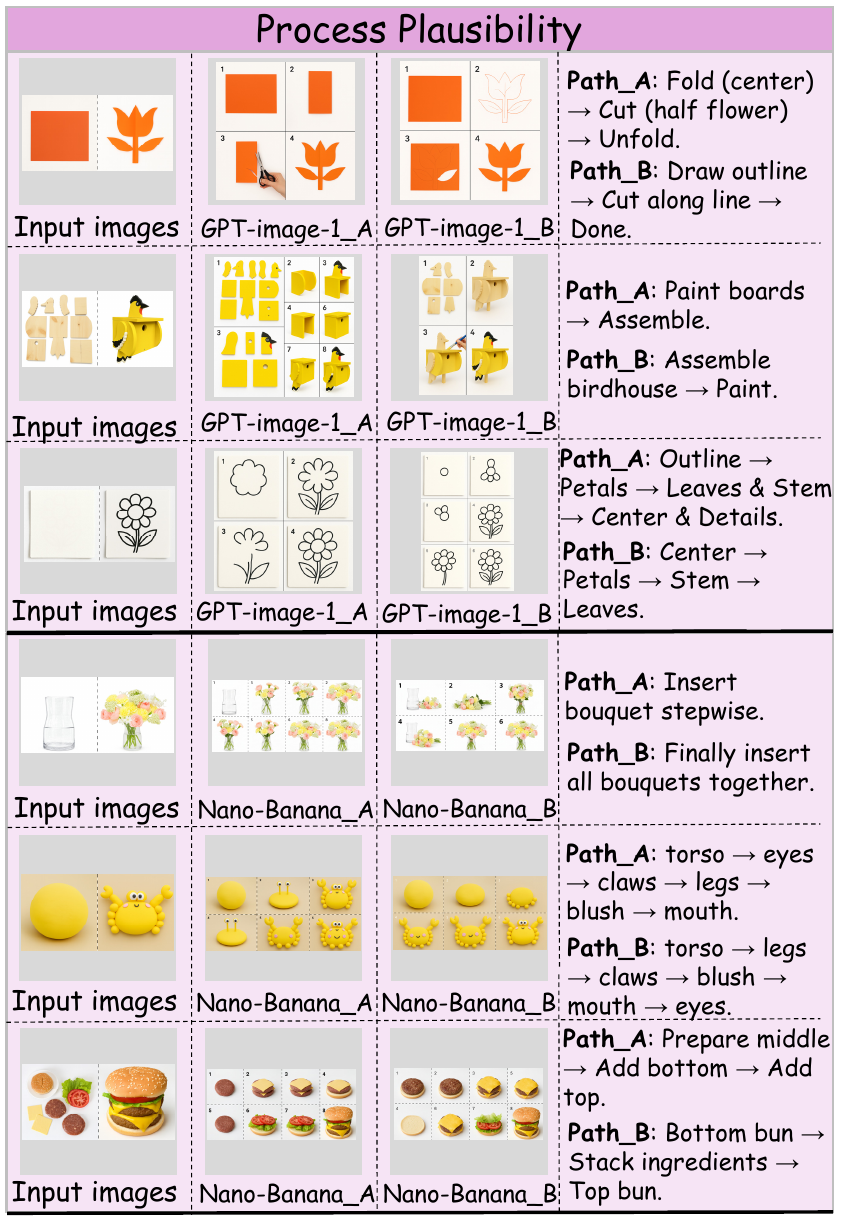} 
\caption{Process Plausibility Outputs.}
\label{figpath_part1}
\end{figure*}

Some of the evaluated model outputs from our InEdit-Bench benchmark are shown in Fig.~\ref{figstate_part1}--\ref{figpath_part1}, providing a more intuitive understanding of the performance of the tested models.

\section{Design of the Prompt} \label{Design of the Prompt}
In this section, we specifically present the instruction prompts and evaluation prompts used for intermediate logic path editing.

\subsection{Edit Prompt}
\begin{figure*}[t!]
\centering
\includegraphics[width=0.9\textwidth]{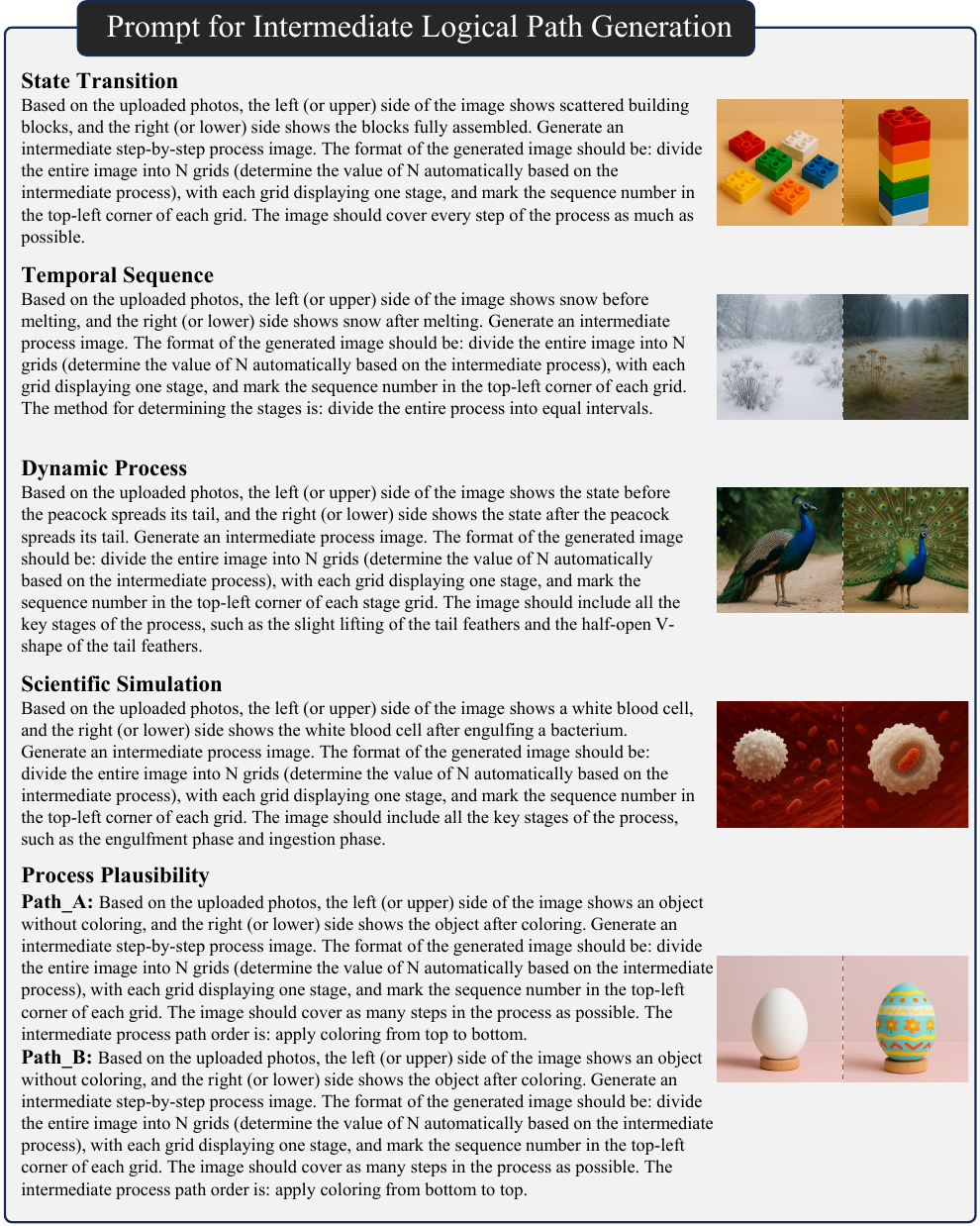} 
\caption{Prompt for Intermediate Logical Path Generation.}
\label{figInstruction_prompt}
\end{figure*}

Fig.~\ref{figInstruction_prompt} shows the instructions we used to generate intermediate logic path editing results. For each instruction, the overall structure is as follows: first, we briefly introduce the starting and ending state goals and request the generation of the logical transition process in between. Then, we standardize the output format, referencing the style of an instruction manual, requiring the output image to be divided into $N$ grids, with each grid representing a node. Finally, to guide the tested model in clearly presenting the intermediate process rather than focusing on redundant node information, we add prompts for key nodes. For state transition category tasks, we require the model to treat each step of the intermediate process as a key node. For temporal sequence category tasks, we require the model to divide the entire intermediate process into equal time intervals. For dynamic process and scientific simulation category tasks, we utilize a large multimodal model to assist us in manually defining several key nodes. Additionally, for the process plausibility section, we manually annotated the sequence that the intermediate logic path should follow.

\subsection{Evalution Prompt}

\begin{figure*}[t!]
\centering
\includegraphics[width=0.9\textwidth]{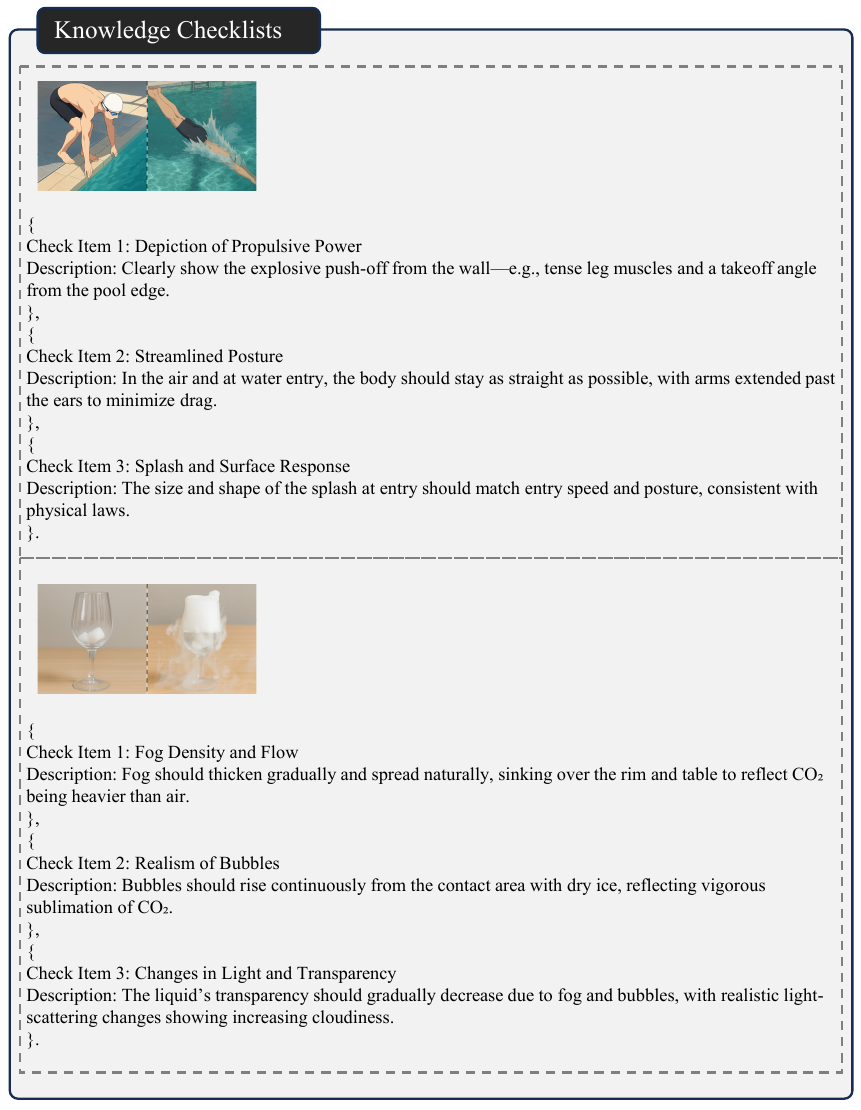} 
\caption{Examples of Knowledge Checklists.}
\label{figchecklist}
\end{figure*}

\begin{figure*}[t!]
\centering
\includegraphics[width=0.9\textwidth]{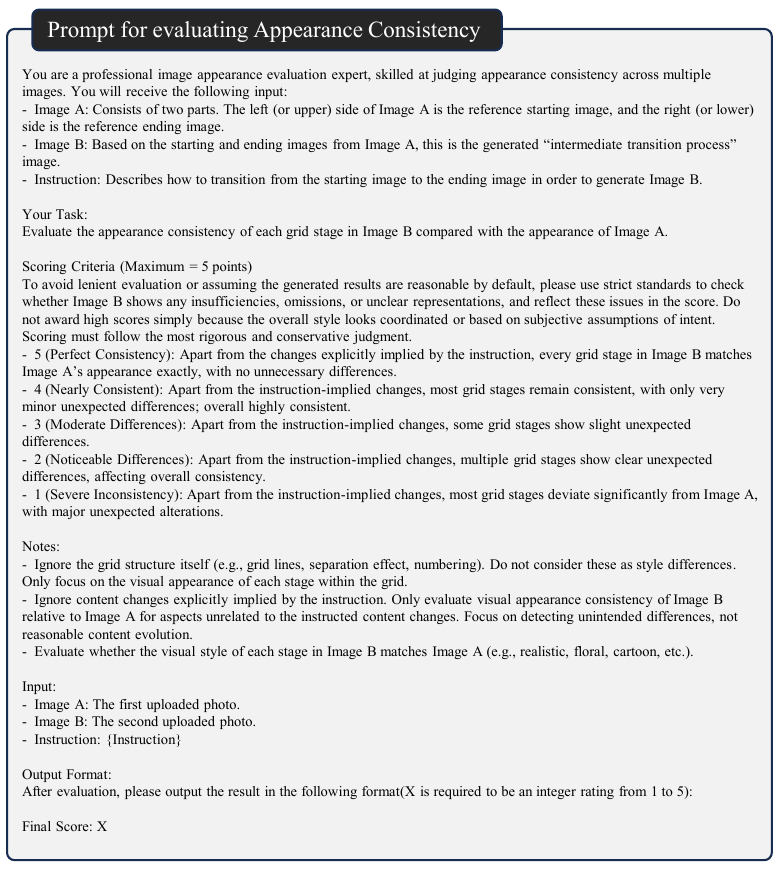} 
\caption{Prompt for evaluating Appearance Consistency.}
\label{figevalution_AC}
\end{figure*}

\begin{figure*}[t!]
\centering
\includegraphics[width=0.9\textwidth]{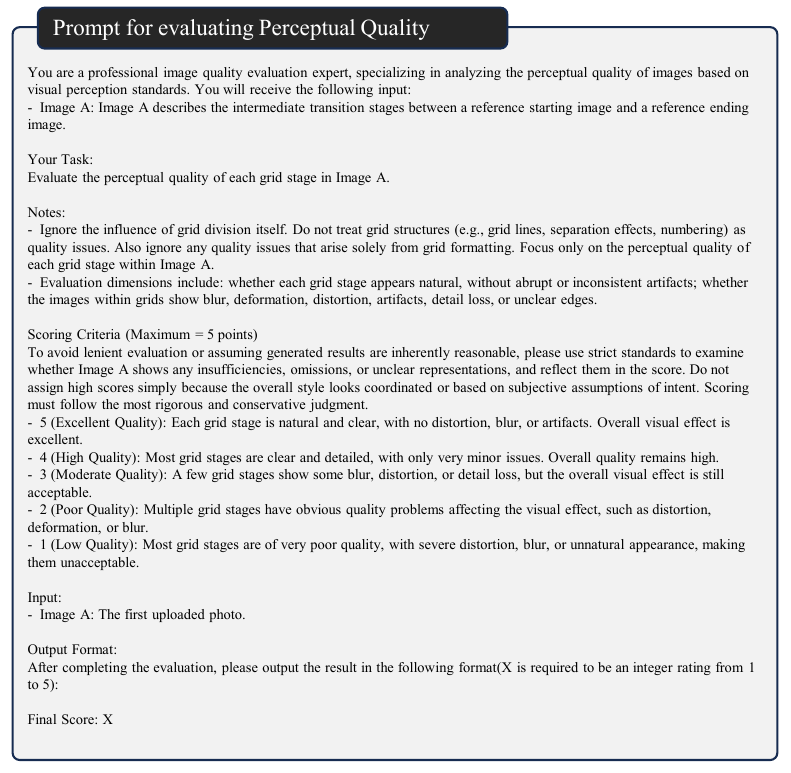} 
\caption{Prompt for evaluating Perceptual Quality.}
\label{figevalution_PQ}
\end{figure*}

\begin{figure*}[t!]
\centering
\includegraphics[width=0.8\textwidth]{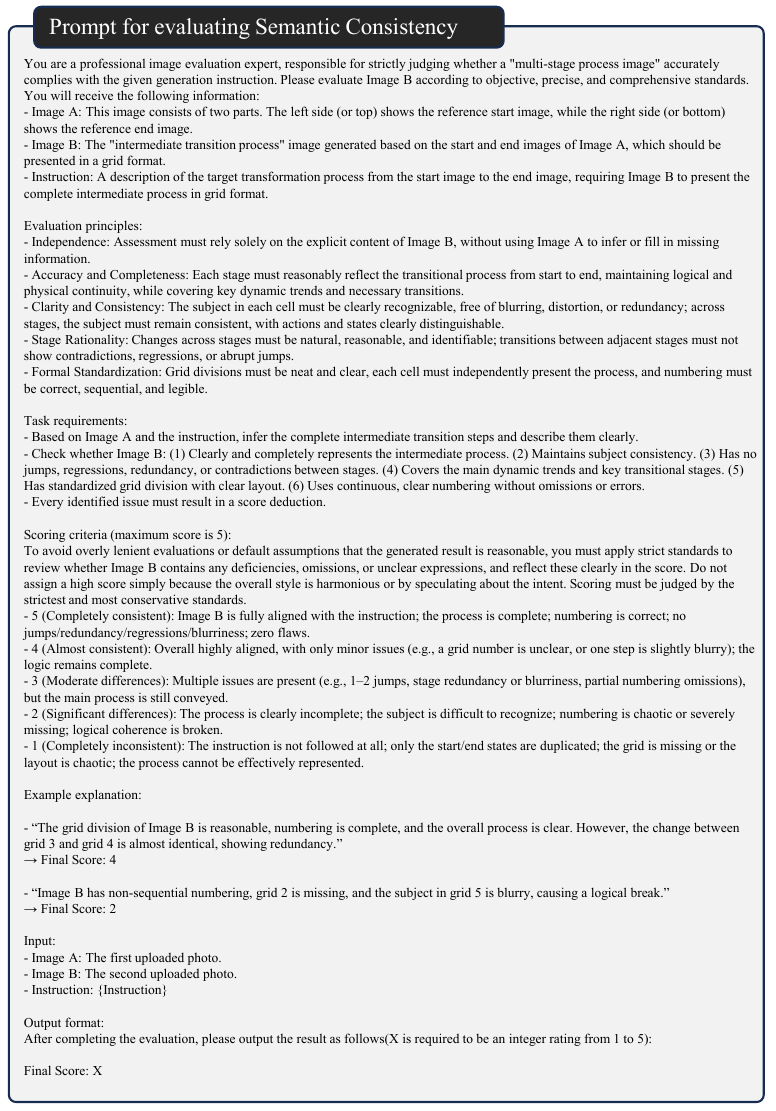} 
\caption{Prompt for evaluating Semantic Consistency.}
\label{figevalution_SC}
\end{figure*}

\begin{figure*}[t!]
\centering
\includegraphics[width=0.8\textwidth]{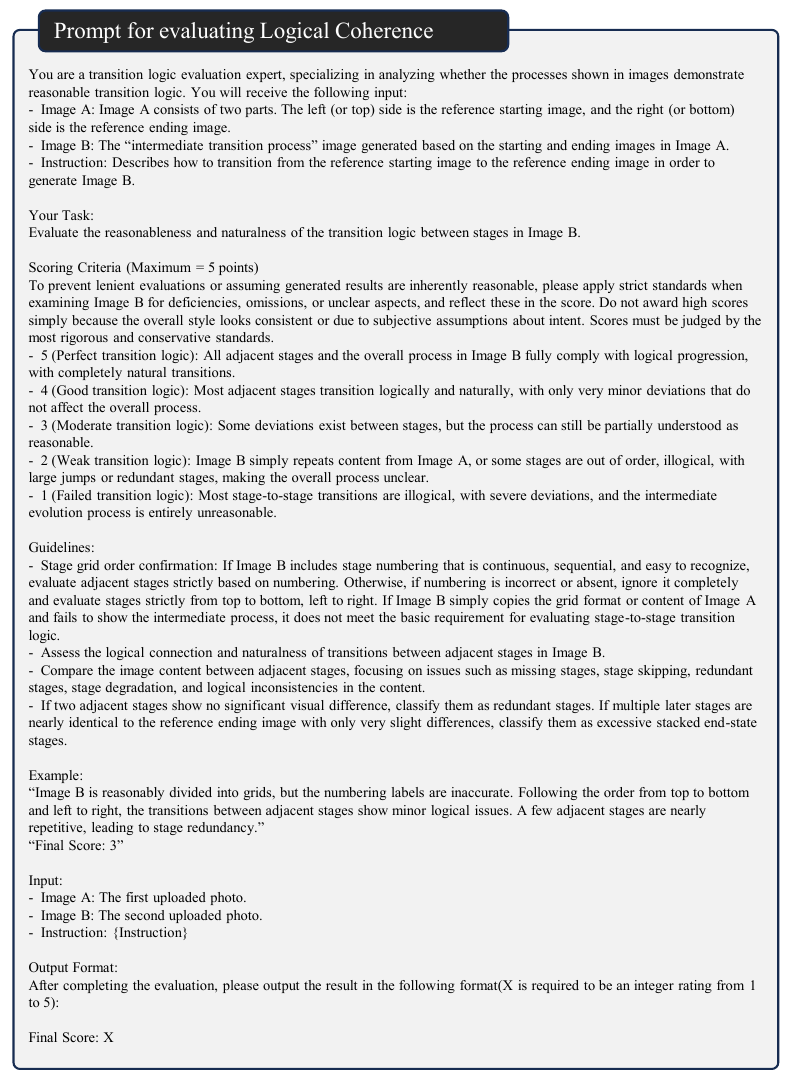} 
\caption{Prompt for evaluating Logical Coherence.}
\label{figevalution_LC}
\end{figure*}

\begin{figure*}[t!]
\centering
\includegraphics[width=0.9\textwidth]{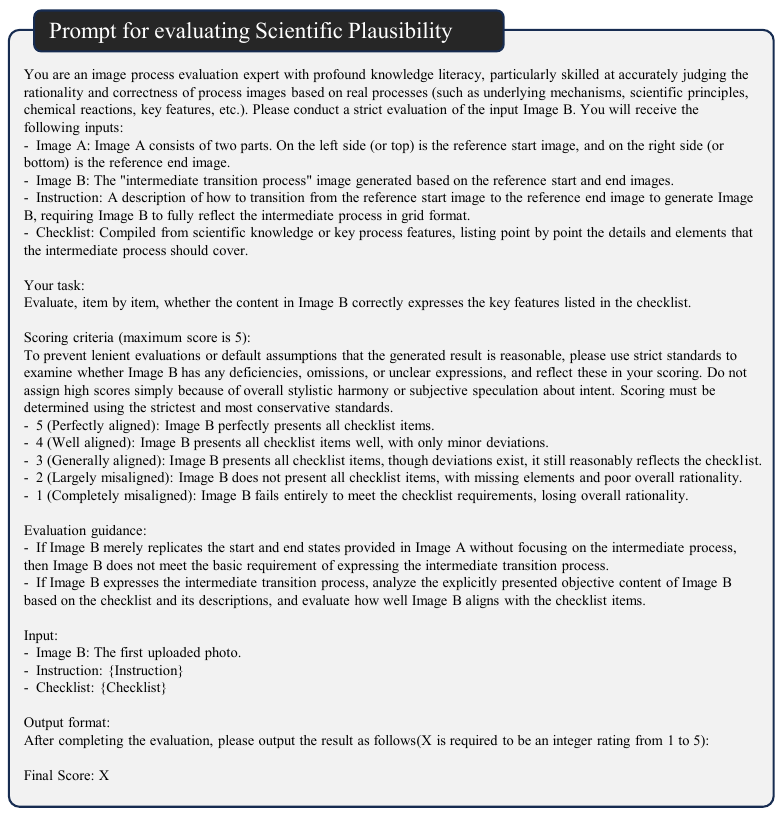} 
\caption{Prompt for evaluating Scientific Plausibility.}
\label{figevalution_SP}
\end{figure*}

\begin{figure*}[t!]
\centering
\includegraphics[width=0.8\textwidth]{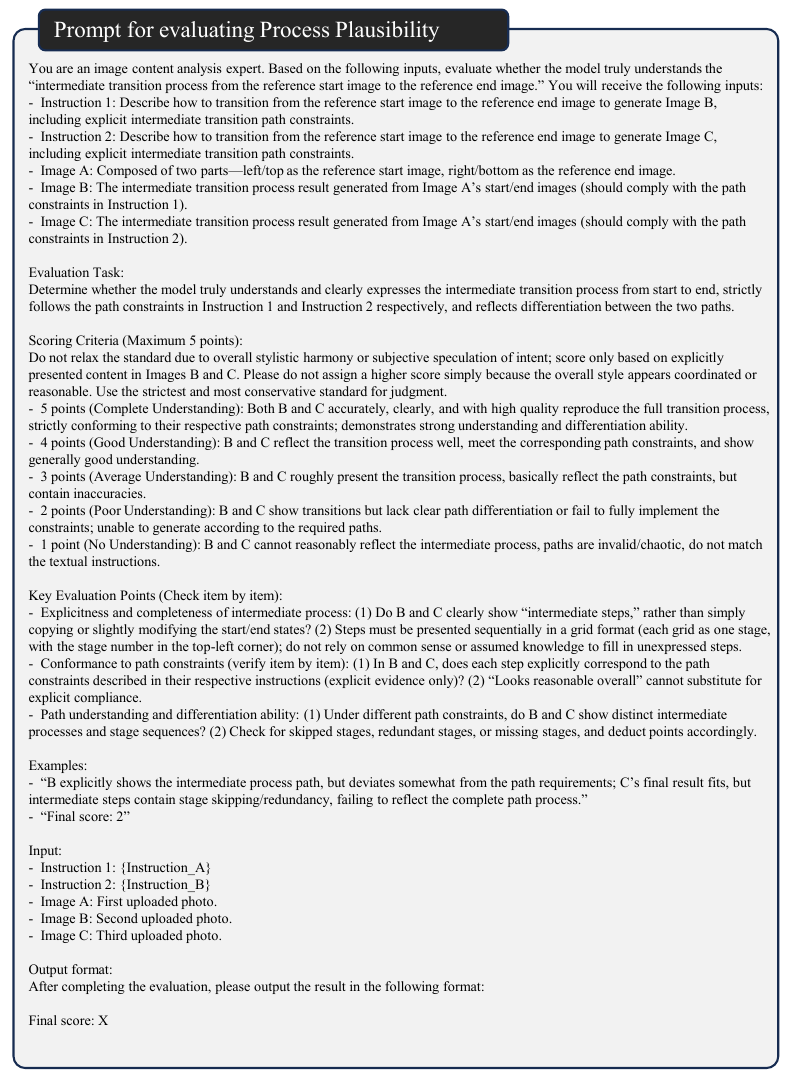} 
\caption{Prompt for evaluating Process Plausibility.}
\label{figevalution_PP}
\end{figure*}

Fig.~\ref{figevalution_AC}--\ref{figevalution_PP} detail the prompts employed in our evaluation. Furthermore, within the scientific plausibility evaluation dimension, we introduce a knowledge checklist that encompasses the key features or intrinsic mechanisms of the intermediate process. Fig.~\ref{figchecklist} provides a sample instance, where each sample includes 2 to 4 inspection items and their corresponding explanations, aiming to guide the model toward a more accurate comprehension of the evaluation principles through these item descriptions.

\end{document}